\documentclass[acmtog]{acmart}

\usepackage{booktabs} 
\usepackage{overpic}
\usepackage{subcaption}
\usepackage{bm}
\usepackage{array}
\usepackage{textcomp}
\usepackage{caption}
\usepackage{xspace}
\usepackage{multirow}
\usepackage{url}
\usepackage{wrapfig}
\usepackage{graphicx}
\renewcommand\footnotetextcopyrightpermission[1]{} 

\author{Songyin Wu}
\email{s_wu975@ucsb.edu}
\orcid{0009-0009-9581-2506}
\affiliation{%
  \institution{University~of~California,~Santa~Barbara}
  \country{USA}
}

\author{Deepak Vembar}
\email{deepak.s.vembar@intel.com}
\orcid{0009-0005-6539-7103}
\affiliation{%
  \institution{Intel Corporation}
  \country{USA}
}

\author{Anton Sochenov}
\email{anton.sochenov@intel.com}
\affiliation{%
  \institution{Intel Corporation}
  \country{USA}
}

\author{Selvakumar Panneer}
\email{selvakumar.panneer@intel.com}
\affiliation{%
  \institution{Intel Corporation}
  \country{USA}
}

\author{Sungye Kim}
\email{sungyekim@gmail.com}
\orcid{0000-0003-0219-2192}
\affiliation{%
  \institution{Intel Corporation (now AMD)}
  \country{USA}
}

\author{Anton Kaplanyan}
\email{anton.kaplanyan@intel.com}
\orcid{0000-0002-8376-6719}
\affiliation{%
  \institution{Intel Corporation}
  \country{USA}
}

\author{Ling-Qi Yan}
\email{lingqi@cs.ucsb.edu}
\orcid{0000-0002-9379-094X}
\affiliation{%
  \institution{University~of~California,~Santa~Barbara}
  \country{USA}
}

\usepackage[ruled]{algorithm2e} 

\SetAlFnt{\small}
\SetAlCapFnt{\small}
\SetAlCapNameFnt{\small}
\SetAlCapHSkip{0pt}

\begin{document}

\title{GFFE: G-buffer Free Frame Extrapolation for Low-latency Real-time Rendering}

\def\LL{{\mathcal{L}}}
\def\bu{{\mathbf{u}}}
\def\bp{{\mathbf{p}}}
\def\bn{{\mathbf{n}}}
\def\bx{{\mathbf{x}}}
\def\by{{\mathbf{y}}}
\def\bo{{\bm{\omega}}}
\def\dd{{\,\mathrm{d}}}

\begin{abstract}
Real-time rendering has been embracing ever-demanding effects, such as ray tracing. However, rendering such effects in high resolution and high frame rate remains challenging. Frame extrapolation methods, which don't introduce additional latency as opposed to frame interpolation methods such as DLSS 3 and FSR 3, boost the frame rate by generating future frames based on previous frames. However, it is a more challenging task because of the lack of information in the disocclusion regions, and recent methods also have a high engine integration cost due to requiring G-buffers as input. We propose a \emph{G-buffer free} frame extrapolation, GFFE, with a novel heuristic framework and an efficient neural network, to plausibly generate new frames in real-time without introducing additional latency. We analyze the motion of dynamic fragments and different types of disocclusions, and design the corresponding modules of the extrapolation block to handle them. After filling disocclusions, a light-weight shading correction network is used to correct shading and improve overall quality. 
GFFE achieves comparable or better results compared to previous interpolation as well as G-buffer-dependent extrapolation methods, with more efficient performance and easier game integration.
\end{abstract}


\begin{teaserfigure}
    \centering
    \begin{overpic}[width=\linewidth]{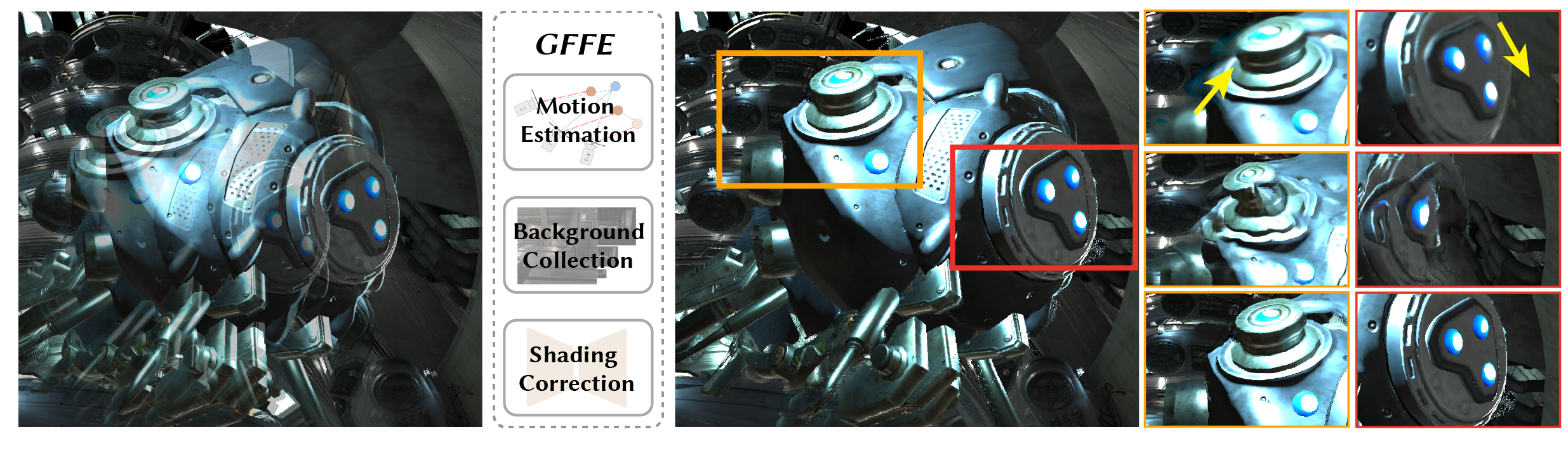}
    \put(2.2, 0){\color{black}\small{Prevous Frames (overlaid for visualization)}}
    \put(53, 0){\color{black}\small{Our Extrapolation}}
    \put(91, 21.5){\color{white}\small{UPR (Interp.)}}
    \put(94, 20){\color{white}\small{43.04ms}}
    \put(88, 12.5){\color{white}\small{DMVFN (}\color{red}\small{Extrap.}\color{white}\small{)}}
    \put(94, 11){\color{white}\small{20.57ms}}
    \put(90, 3.22){\color{white}\small{Ours (}\color{red}\small{Extrap.}\color{white}\small{)}}
    \put(94.5, 1.82){\color{yellow}\small{6.62ms}}
    \end{overpic}
    \caption{We propose a \emph{G-buffer free} frame extrapolation framework, GFFE, which introduces no additional latency (unlike interpolation methods) and eliminates the need for additional G-buffers generation of extrapolated frames. Our framework shows better visual quality than previous frame extrapolation method DMVFN~\cite{hu2023dynamic} and interpolation method UPR~\cite{jin2023unified} with better performance.}
    \label{fig:teaser}
\end{teaserfigure}
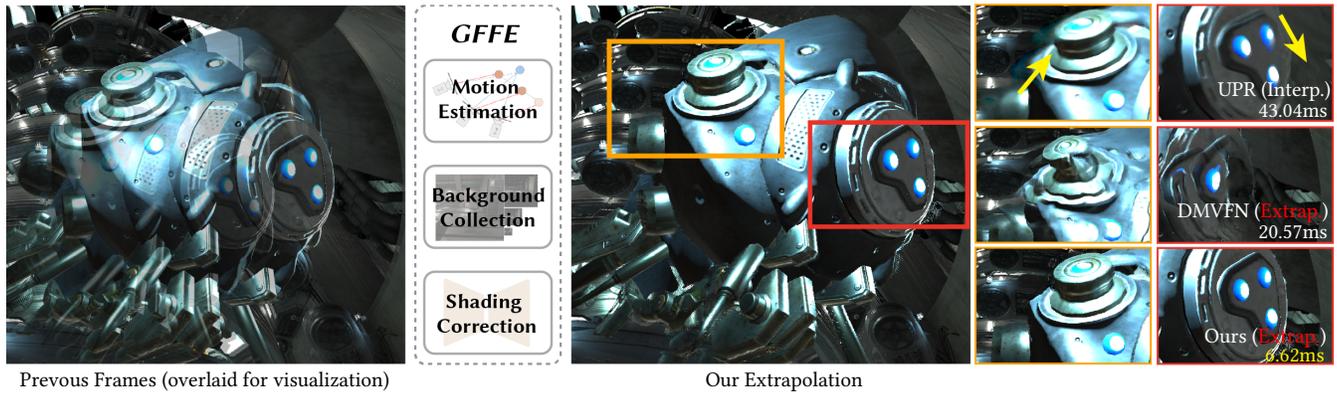

\maketitle

\section{Introduction}
Real-time rendering has advanced significantly in recent years to create more realistic and interactive  environments, including the recent trend for real-time path tracing effects in games. Usually, high quality and high frame rates are required for games or virtual reality applications in order to provide a good user experience. However, the cost of rendering such high quality frames is expensive even for the most powerful graphics hardware - naively rendering all frames is not always possible under fixed compute and power budgets. Therefore, in addition to methods that accelerate frame rendering, approaches such as frame super resolution and generation~\cite{guo2021extranet,wu2023extrass,wu2023lmv,xiao2020neural, guo2022classifier,xess, fsr} are usually implemented in a separate post-processing pass to provide the best quality output within given compute budgets.  

Frame generation is one technique that can be used to increase the frame rate for smoother and jitter-free experience. Frame interpolation, including proprietary products DLSS 3~\cite{dlss3} and FSR 3~\cite{fsr3}
and research works~\cite{briedis2021neural,briedis2023kernel,jin2023unified,kong2022ifrnet} try to generate new frames between two rendered frames. These methods increase the key-press-to-display latency of the rendering process since the generated frames rely on availability of both the previous and the next frame.

Frame extrapolation, on the other hand, generates new frames based solely on previous frames, and does not introduce additional latency to the rendering process. However, it is a more difficult task and usually generates inferior results due to the missing information from the future frames. Many existing methods, including ExtraNet~\cite{guo2021extranet}, LMV~\cite{wu2023lmv} and ExtraSS~\cite{wu2023extrass}, use G-buffers of generated frames to guide the generation of corresponding final frames. Game-generated G-buffers are not always easily available and the cost of obtaining them from various rendering pipelines is not negligible. 
Other video extrapolation methods~\cite{hu2023dynamic} do not require G-buffers to generate color frames, however they usually have inferior quality and performance under real-time rendering settings.

Existing methods have shown abilities to generate new frames, but they either introduce latency or require additional G-buffers. Motivated by these problems, we propose a novel method that can generate new frames without introducing latency or requiring G-buffers. 
Our insight is the missing information of extrapolated frames can be approximately retrieved from previous frames, which are usually discarded in the rendering pipeline. Additionally, the motion of fragments can be plausibly estimated from history frames, so there is no need to render G-buffers for extrapolated frames. 

Based on these observations, we propose a G-buffer-free extrapolation framework.  First, it uses a heuristic motion estimation method to eliminate the requirement of rendering motion vectors for extrapolated frames. Then, to handle disocclusions in the extrapolated frames, we introduce a background collection module and adaptive rendering window. Lastly, we use a light-weight neural network to further improve the shading and shadow consistency. 

We evaluate our framework on various scenes in Unreal Engine~\cite{unrealengine} with different types of effects including glossy and translucent materials, complex geometry, and dynamic objects to demonstrate our quality, performance, and robustness. Our method generates smooth and plausible results from 30 FPS to 60FPS. It shows superior quality to G-buffer free extrapolation baseline, and comparable results with G-buffer-dependent baseline and interpolation baselines, with better visual quality and performance.

\section{Related Work}
\label{sec:related_work}

\begin{figure}
    \centering
    \begin{overpic}[width=\columnwidth]{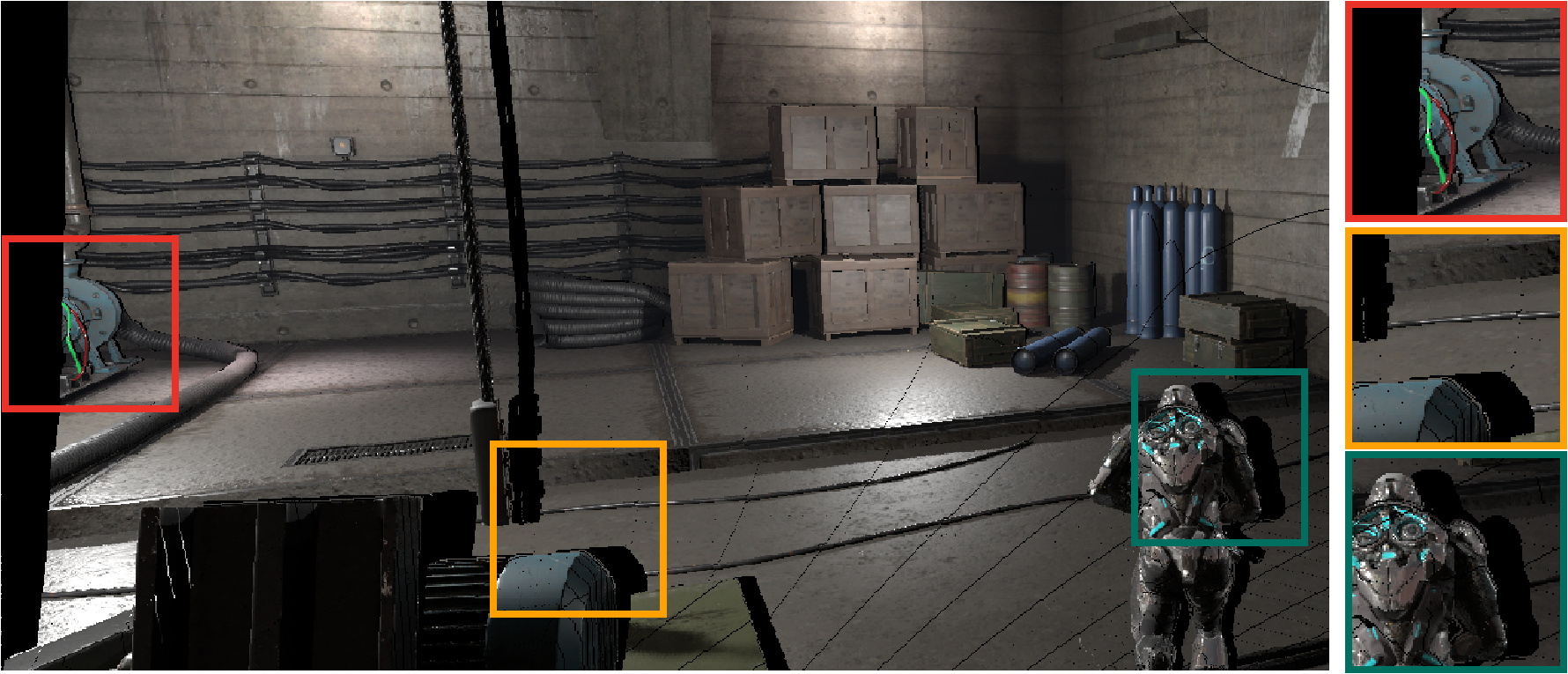}
    \end{overpic}
    \caption{An extrapolated frame by directly projecting fragments from previous rendered frame. The right column shows three types of disocclusions: \textit{out-of-screen disocclusion}, \textit{static disocclusion} and \textit{dynamic disocclusion} from top to bottom. The thin black lines splatted in the image are due to forward warping.}
    \label{fig:disocclusions}
\end{figure}

\subsection{Warping and hole filling}
Warping has been used in real time rendering for many years to improve the quality and performance. Mark et al.~\shortcite{mark1997post} proposed a 3D warping method to warp the frame to new frames as a post processing step. But it is difficult for disocclusion areas since such informaiton is not available, where a hole filling algorithm is needed. Didyk et al.~\shortcite{didyk2010perceptually} employed additional blur operations to the warped frames to reduce the artifacts for disocclusion areas. Similar to \citet{didyk2010perceptually}, Schollmeyer et al.~\shortcite{schollmeyer2017efficient} proposed a hole filling method to fill the disocclusion areas by low pass filtering of them to reduce the artifacts. These methods bring blurring artifacts to the extrapolated frames instead of generating actual details, which is not suitable for modern real-time rendering. Later, Reinert et al.~\shortcite{reinert2016proxy} builds geometry proxies to fill the disocclusion areas but requires pre-computed geometry information and still in low quality since it uses low poly geometries. \cite{zeng2021temporally,wu2023extrass} use G-buffers to guide the hole filling process by reusing spatial neighbors' information. However, the G-buffers are not always available in real-time rendering, which limits the usage of these methods.

Besides single frame warping methods which cannot retrieve valid information in disocclusions, there are also some bidirectional methods trying to warp frames from both previous and future frames. Andreev~\shortcite{andreev2010real} uses half motion vectors to warp both previous frame and future frame to the current frame to increase the frame rate. Yang et al.\shortcite{yang2011image} uses an iterative way to find the correspondence from previous and future frames to the current frame. These methods, although fill the disocclusion areas better, introduce additional key-press-to-display latency since new frames reply on future frames and usually the quality is not good enough including lagging shadow and shadings.

\subsection{Frame Interpolation}
Besides pure warping based methods, there are several frame interpolation methods with neural networks achieving better quality. \cite{briedis2021neural,briedis2023kernel} propose using optical flows or kernel prediction neural network to generate intermediate frames by only given corresponding G-buffers. Although the quality looks promising, these techniques are used for offline rendering, which is difficult to be applied in real time rendering due to low performance. 
Video interpolation methods~\cite{huang2022real,jin2023unified,kong2022ifrnet, bao2019depth} also generate plausible intermediate frames with neural networks but usually with blurrier results and worse performance since these methods are not designed for rendering pipeline. 
Commercial solutions including DLSS 3~\cite{dlss3} and FSR 3~\cite{fsr3} are also proposed to boost frame rate in games but the details of their methods are not released. They are usually running under very high frame rate so that the artifacts become less noticeable. Offline frame interpolatin methods~\cite{zhou2022exploring, zhang2023extracting, reda2022film} take more than 100ms per frame which is impratical in real-time rendering engine.
Nevertheless, frame interpolation methods bring more key-press-to-display latency, making users feel lagging when interacting with the scene, and this becomes more severe when input frame rate is low such as boosting 30 FPS to 60 FPS.

\subsection{Frame Extrapolation}
To avoid the extra latency introduced by frame interpolation while increasing frame rate, frame extrapolation methods have been studies these years to generate new frames only based on history frames. ExtraNet~\cite{guo2021extranet} uses occlusion motion vectors~\cite{zeng2021temporally} with a neural network to handle both disocclusion areas and lagging shadow and shadings. Learnable motion vector~\cite{wu2023lmv} proposes a recurrent framework to optimize motion vectors so that they can handle the motion of shadings and disocclusion areas. ExtraSS~\cite{wu2023extrass} uses G-buffers to guide the extrapolation process and uses a flow-based neural network to fix the shading errors. All of these methods require generation of G-buffers for extrapolated frames, which is not always the case in real-time rendering of different engines and platforms such as mobile, cloud gaming and some forward rendering engines. Concurrent work \citet{yang2024mobfgsr} uses simple warping and hole filling method for extrapolation but fails with large disocclusions and does not consider shading's motion. 

Video extrapolation methods~\cite{hu2023dynamic}, although do not require G-buffers for extrapolated frames, usually yield much worse quality and performance, which are usually not suitable for real-time rendering. \cite{li2022future} uses optical flow to predict the future frames but a reshading process is needed for refining extrapolated frames, which is different from our settings.

\section{Motivation}

\begin{table}
    \begin{center}

    \setlength{\tabcolsep}{0.9mm}{
    \renewcommand{\arraystretch}{1.1} {
        \caption{Features and challenges of different frame generation methods.}
        \begin{tabular}{ c | c c c }

        & G-buf Free & G-buf Dependent & G-buf Free \\
        & Interp. & Extrap. & Extrap. \\ \hline
         Low latency &  & \checkmark & \checkmark  \\
         No-extra G-buffers& \checkmark & & \checkmark \\ \hline
          Motion Est.& \checkmark &  & \checkmark \\
         Disocclusion &  & \checkmark &  \checkmark\\
         Non-geo Tracking  & \checkmark & \checkmark & \checkmark \\ 

        \end{tabular}
        \label{tlb:background}
        }
    }
    \end{center}
\end{table}

\subsection{Problem formulation and design choices}
\label{sec:problem_formulation}
Our G-buffer free extrapolation framework aims to extrapolate new frames to increase the presented frame rate without dependence on G-buffers for extrapolated frames and additional latency. 
Note that we use the term "G-buffer free" to refer to the absence of G-buffers for extrapolated frames only. The depth buffer and motion vectors for rendered frames are used since they are usually readily available in the rendering engine without additional cost.
Unlike previous G-buffer based extrapolation methods~\cite{wu2023extrass,wu2023lmv}, the G-buffers for extrapolated frames are not available under our setting and some types of G-buffers including albedo, normal and roughness are not available even for rendered frames in forward and in some cases deferred renderers.

We formulate our problem as follows, given a sequence of rendered frames $\{I_t\}$ with their corresponding depth buffer $\{D_t\}$ and motion vectors $\{V_t\}$, our framework generates new frames $\{\bar{I}_{t+\alpha}\}$ with their corresponding depth buffer $\{\bar{D}_{t+\alpha}\}$ and motion vectors $\{\bar{V}_{t+\alpha}\}$, where $\alpha$ depends on the number of frames we want to generate for every rendered frame. 

In addition to our G-buffer free frame extrapolation, there are two other methods commonly used: G-buffer free frame \textbf{interpolation} and G-buffer \textbf{dependent} frame extrapolation. The features of these three types of methods are shown in Table~\ref{tlb:background}.

\paragraph{Latency} Frame interpolation methods are widely used and have demonstrated good quality as it is easier to find correspondence in either previous or latter frames. The main disadvantage of interpolation methods is the additional latency introduced. As analyzed in previous works~\cite{wu2023extrass,wu2023lmv,guo2021extranet}, the latency of interpolation methods is increased by at least one rendering time interval, which is even higher than the original latency without the frame interpolation method. This leads to worse user experience especially when the low latency is required such as competitive games~\cite{kim2020post} and VR applications. Although mitigation techniques such as NVIDIA Reflex can be used to decrease the latency, they still cannot fully eliminate it. 

\paragraph{G-buffers} To avoid introducing additional latency, frame extrapolation methods have been proposed~\cite{wu2023extrass,guo2021extranet,wu2023lmv}. Since it is a more challenging task to achieve similar quality compared to frame interpolation methods, various types of G-buffers from extrapolated frames are required. However, it is not always practical due to the different engine types (forward rendering engine) and G-buffer generation cost. More discussion about this is included in the appendix.

Besides, we also consider our G-buffer free frame extrapolation framework for possible future applications, especially low-latency streaming and cloud gaming on various low end devices. In order to provide immediate response to the user inputs, the frame extrapolation should be done in the client side, where scene information is not available to generate G-buffers. Therefore, our G-buffer free frame extrapolation framework is more suitable for such applications than frame interpolation or G-buffer dependent extrapolation.

\subsection{Challenges}
\label{sec:challenge}
\subsubsection{Motion estimation}

Our method works under the assumption that the rendering engine doesn't generate any G-buffers for extrapolated frames. Therefore, unlike previous G-buffer dependent extrapolation methods~\cite{wu2023extrass, guo2021extranet, wu2023lmv}, the motion from rendered frame $I_t$ to extrapolated frame $\bar{I}_{t+\alpha}$ is unknown. Previous warping methods~\cite{mark1997post,lee2018iterative,bowles2012iterative} work either only on static scenes or where the objects' motion is given, so no motion estimation is needed. 

Motion estimation is challenging since the motion of the objects in the game can be arbitrarily complex. 
Frame interpolation methods~\cite{jin2023unified,kong2022ifrnet} use neural networks to predict the motion between two rendered frames but they are usually slow and unstable sometimes. 

Instead of using heavy and slow neural networks to predict the motion, we use a heuristic motion approximation method to estimate the motion for each dynamic fragment for extrapolated frames. Our goal is to estimate a plausible motion in order to achieve smooth transitions in continuous frames - we do not expect perfectly estimated motion since the future frame's motion can be arbitrary.

\subsubsection{Disocclusions}
\label{sec:disocclusion}

Disocclusions, as shown in Fig.~\ref{fig:disocclusions}, are areas that are not shown in the previous frame but visible in the current frame.
They are challenging to handle with the frame extrapolation approach due to the lack of information from the next rendered frame, which is used in frame interpolation frameworks. Previous frame extrapolation methods used G-buffers for the extrapolated frame. Although they are not shaded, they provide sufficient information to fill disocclusions. However, under our settings, there is no such information in either the previous frame or G-buffers, which makes it much more difficult to recover this information. 

To better understand and deal with the disocclusions, we categorize them into three types: 
\textbf{(1) Out-of-screen disocclusion:} Pixels that are shown in the current frame but are not in the screen space of the previous frame. This type of disoccluded areas are caused by the camera's motion.
\textbf{(2) Static disocclusion:} Pixels that are shown in the current frame but are not shown in the previous frame due to occlusion from static occluders. These pixels are static in the two consecutive frames and in the screen space of the previous frames. They become visible in the current frame due to the change of the camera's position.
\textbf{(3) Dynamic disocclusion:} Similar to the static disocclusion, the only difference is that the occluders are dynamic. These areas are usually caused by the motion of the occluders instead of the camera.

Simply using a neural network to fill the disocclusions causes severe artifacts as we don't have any information in those areas and the size\footnote{The size of disocclusions depends on the frame rate and objects' motion speed. We target 30 FPS inputs which usually has noticeable disocclusions} of the disocclusions areas is usually not small as shown in Fig.~\ref{fig:disocclusions}. Our proposed method uses history information with efficient adaptive rendering windows to handle the disocclusions more plausibly.

\subsubsection{Non-geometric motion tracking}
\label{sec:shading}
Frame generation methods usually reuse temporal information, and try to find corresponding pixels in existing frames. However, such correspondence computation is not always accurate. The color in the rendered frames is the combination of lighting information with materials' properties, which may have different directions of motion. Rendered motion vectors only capture the motion of the geometry, but not the motion of the lighting information. Only considering the geometries' motion like \citet{yang2024mobfgsr} will cause lagging in shading and shadows~\cite{guo2021extranet, wu2023extrass, wu2023lmv}. For example, shadows move at 30 FPS and other objects move at 60 FPS.
These shadows and reflections, although they contribute a small portion to common metrics such as PSNR and SSIM, are crucial for the visual quality.
Therefore, a module for tracking such motion is necessary to maintain a high frame rate in all areas and provide smooth transition between frames. To address this, we designed our shading correction network to fix these issues.

\section{Method}

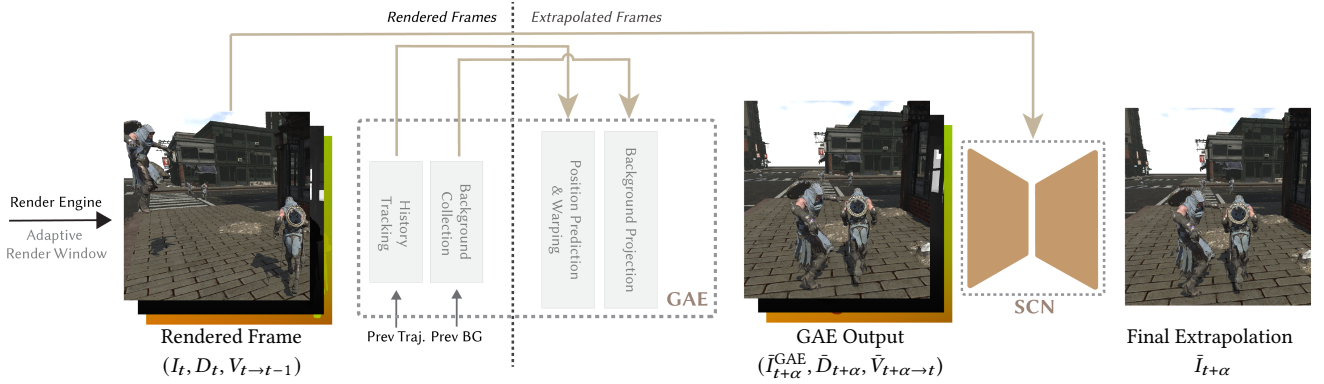
\begin{figure*}
    \centering
    \begin{overpic}[width=\linewidth]{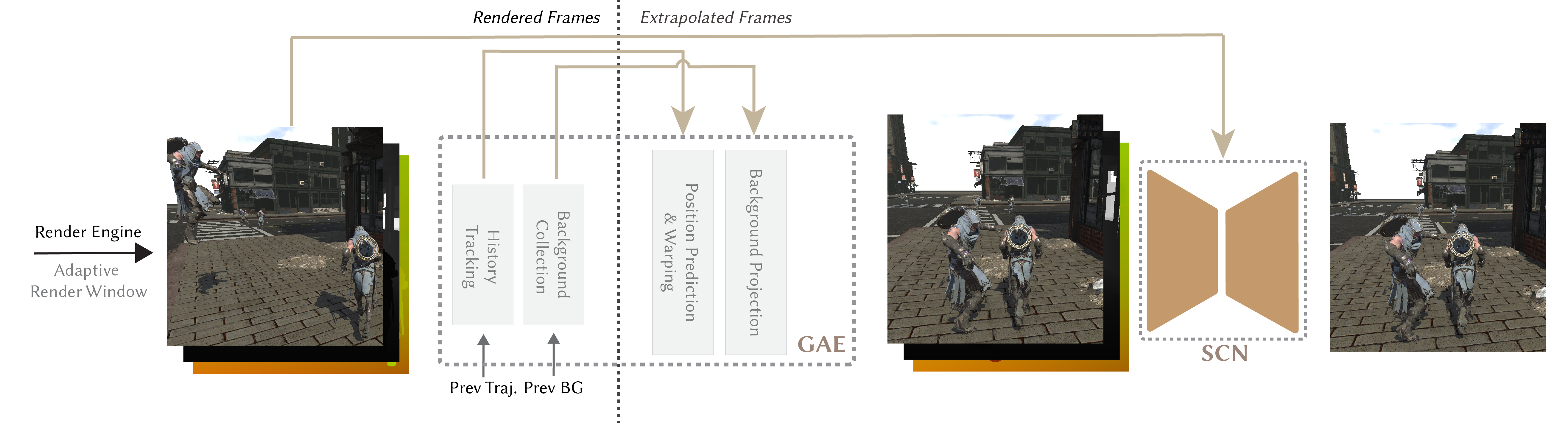}
    \put(13.4, 1.7){\color{black}\small{Rendered Frame}}
    \put(13.4, -0.4){\color{black}\small{$(I_t, D_t, V_{t\rightarrow t-1})$}}

    \put(60.5, 1.7){\color{black}\small{GAE Output}}
    \put(57.5, -0.4){\color{black}\small{$(\bar{I}_{t+\alpha}^{\text{GAE}}, \bar{D}_{t+\alpha}, \bar{V}_{t+\alpha\rightarrow t}$)}}

    \put(85, 1.7){\color{black}\small{Final Extrapolation}}
    \put(90, -0.4){\color{black}\small{$\bar{I}_{t+\alpha}$}}
    
    \end{overpic}
    \caption{Our method generates an extrapolated frame $\bar{I}_{t+\alpha}$ from the rendered frame $I_t$ and history frames. 
    The left part shows the process of rendered frames including adaptive rendering window , history tracking and background collection, which are prepared for extrapolated frames. The right part shows the process of extrapolating a frame, including geometry aware extrapolation (GAE) and shading correction network (SCN).
    The depth and motion vectors in extrapolated frames are generated in our framework instead of rendering engine, which can be used for additional post-processing.}
    \label{fig:overview}
\end{figure*}

The overview of our framework is shown in Fig.~\ref{fig:overview}. Challenges mentioned in sec.~\ref{sec:challenge} are addressed by our different modules: motion estimation (sec.~\ref{sec:motion_estimation_warping}), background collection for static and dynamic disocclusions (sec.~\ref{sec:background_collection}), adaptive rendering window for out-of-screen disocclusions (sec.~\ref{sec:adaptive_rendering_windows}) and shading correction network for non-geometric motion tracking (sec.~\ref{sec:shading_correction}).
We detail each component in the following sections.

\subsection{Motion estimation}
\label{sec:motion_estimation_warping}

Frame extrapolation usually re-uses history frames information to generate new frames.
Existing extrapolation methods~\cite{wu2023extrass, wu2023lmv,guo2021extranet} use motion vectors from rendering engines to find the corresponding pixels in the previous frame. These motion vectors are accurate but require full rasterization pipeline for extrapolated frames. We propose a motion estimation module to efficiently predict the motion of fragments for extrapolated frames.

The motion estimation module consists of three parts: history tracking, position estimation and warping. It first collects the history trajectory in the world space, and then use it to estimate the next world position for extrapolated frame. After that, a warping process is applied to warp fragments to extrapolated frames.

\paragraph{History tracking} History tracking happens in rendered frames, which calculate the history trajectory of each fragment in the world space. 
In high level, our history tracking algorithm works recurrently for each rendered frame to generate $k$ history world position $\{P_i[x]\}$ for each pixel, where current trajectory is updated from corresponded previous trajectory. In order to avoid incorrect correspondences due to disocclusions, we designed a static test algorithm by comparing previous screen space of each pixels by using motion vectors and view projection matrix. If the distance is small than a threshold, the pixel is set as a static pixel, to avoid calculating incorrect history trajectory.  The details of algorithm is shown in appendix Algo.~\ref{alg:motion_fixing}.

\begin{figure}
    \centering
    \begin{overpic}[width=0.9\columnwidth]{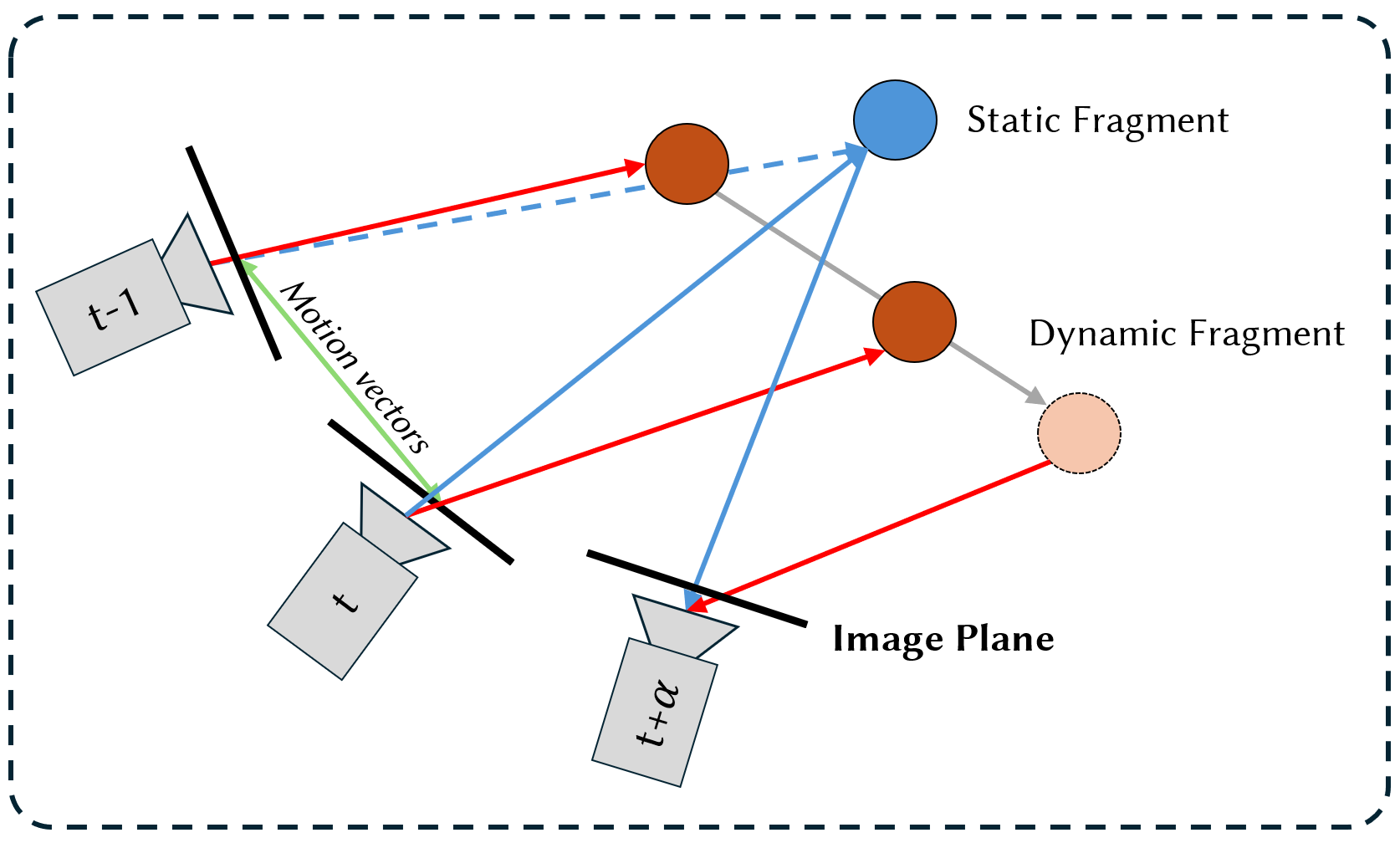}
    \end{overpic}
    \caption{Our motion estimation module tracks history trajectories and estimate next world positions based on history trajectory. }
    \label{fig:forwardMotion}
\end{figure}

\paragraph{Position estimation} History tracking provides history world position $\{P_i[x]\}$ for each pixel. For extrapolated frames, let $\alpha$ be the extrapolation factor which is calculated by $\alpha = \frac{j}{n+1}$, where $n$ is the number of extrapolated frames per rendered frame, and $j$ refers to the $j$-th extrapolated frame for a rendered frame. As shown in Fig.~\ref{fig:forwardMotion}, the next position $NP_{t\rightarrow t+\alpha}$ is estimated by calculating the linear motion of last two position in the trajectory
\begin{equation}
    NP_{t\rightarrow t+\alpha}[x] = \alpha (P_0[x] - P_1[x]) + P_0[x]
\end{equation}

Unlike calculating the motion in the images where linear motions are not reliable due to perspective project, camera rotation and etc., linear motion assumption in the world space efficiently generates plausible next world positions. High order polynomials could be used here but leads to worse results. Please refer to ablation studies for more analysis.

\paragraph{Warping} After calculating the next world position, each fragment is projected to the extrapolated frame based on the camera view projection matrix. For multiple fragments projected into the same pixel, we compare the depth value for the projected pixels and keep the fragment with the smallest depth value using atomic operations. Although there are several works\cite{lee2018iterative, bowles2012iterative} with better ways of warping/projection, our warping method is efficient and simple, which is already sufficient for our pipeline.

\subsection{Hierarchical Background Collection}
\label{sec:background_collection}

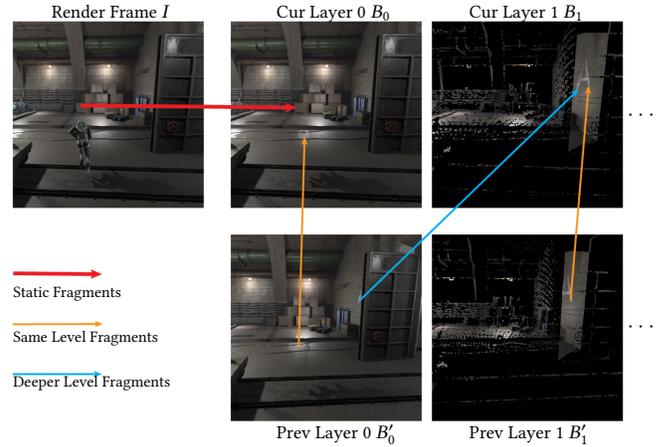
\begin{figure}
    \centering

    \begin{overpic}[width=\columnwidth]{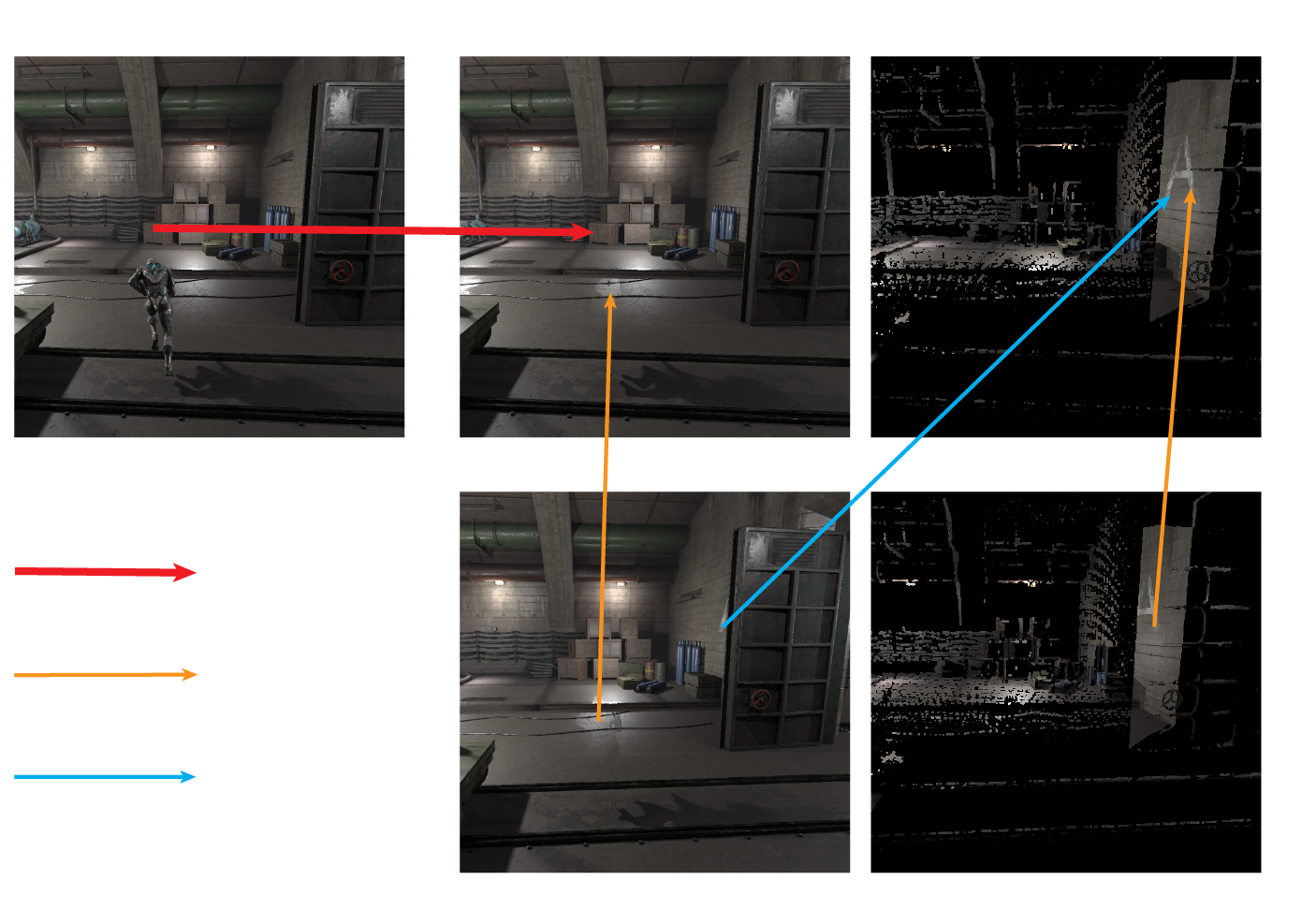}
    \put(7, 65.5){\color{black}\footnotesize{Render Frame $I$}}
    \put(42, 65.5){\color{black}\footnotesize{Cur Layer 0 $B_0$}}
    \put(72, 65.5){\color{black}\footnotesize{Cur Layer 1 $B_1$}}
    \put(42, 0){\color{black}\footnotesize{Prev Layer 0 $B'_0$}}
    \put(72, 0){\color{black}\footnotesize{Prev Layer 1 $B'_1$}}
    \put(1, 22){\color{black}\scriptsize{Static Fragments}}
    \put(1, 15){\color{black}\scriptsize{Same Level Fragments}}
    \put(1, 8){\color{black}\scriptsize{Deeper Level Fragments}}
    \put(96.5, 49.5){\color{black}\small{$\cdots$}}
    \put(96.5, 16.5){\color{black}\small{$\cdots$}}
    \end{overpic}
    \caption{The process of hierarchical background collection. Top row is current rendered frame and updated background buffers, and the bottom row is previous background buffers. Different color arrows show different conditions when updating the background buffers. The size of deeper layers (Layer 1) is only $1/4$ as the previous layer (Layer 0).}
    \label{fig:bgcollection}
\end{figure}

With estimated motion and warping in sec.~\ref{sec:motion_estimation_warping}, an initial extrapolated frame is generated but with invalid regions caused by disocclusions as analyzed in sec.~\ref{sec:disocclusion}, where we use two modules to handle them.
One insight is that static and dynamic disocclusions, although are invisible in the previous frame, may showed in long history frames before. However, naively storing more history frames are impractical due to memory limits and matching the correspondences are also time consuming. Inspired by this, we propose a hierarchical background collection module, to efficiently collect useful information from history frames to fill disocclusions.

This module contains two parts: a background collection for rendered frames to maintain a background buffer $\{B_l\}$ to collect the fragments of rendered frames as well as the fragments behind it without additional rendering cost, and a background projection process for extrapolated frames to fill disocclusions.

\paragraph{Background collection} Fig.~\ref{fig:bgcollection} shows the process of backrgound colleciton. The background buffer $B$ contains $L$ levels with a pair of color buffer and depth buffer for each level, denoting as $B = \{B_l\}$, and the size of deeper level is only $1/4$ to the previous level. Let $(I_t, D_t)$ be the current rendered frame, and $B' = \{B'_l\}$ be the previous background buffer. The static fragments of rendered frame $(I_t, D_t)$ are filled into the first layer of updated background buffer $B_0$. For each level $l$ in the previously collected background $B'_l$, there two conditions to update the current background buffer $B_l$:
\begin{itemize}
    \item \textbf{Case 1 (same level fragments): } If the corresponding position in the same level $B_l[x']$ is invalid, $B'_l[x]$ is used for filling $B_l[x']$.
    \item \textbf{Case 2 (deeper level fragments): } If the corresponding position in the same level $B_l[x']$ is already valid, and the depth value of $B'_l[x]$ is larger than it, $B'_l[x]$ is used for next level $B_{l+1}[x']$, meaning the deeper fragments of the current layer. If multiple fragments are projected into the same pixel, we keep the fragment with the smallest depth value that satisfies the condition.
\end{itemize}

Each level represents a layer of geometries in the scene and the higher level contains the deeper fragments. Hence, we can keep track the occluded fragments by updating the background from level $0$ to $L$. 

\paragraph{Background projection} For extrapolated frames, the collected background buffers are projected to the world space and then back to the extrapolated frames. We only fill the invalid regions of disocclusions in the extrapolated frames.

\subsection{Adaptive rendering window}
\label{sec:adaptive_rendering_windows}

\begin{figure}
    \centering
    \begin{overpic}[width=\columnwidth]{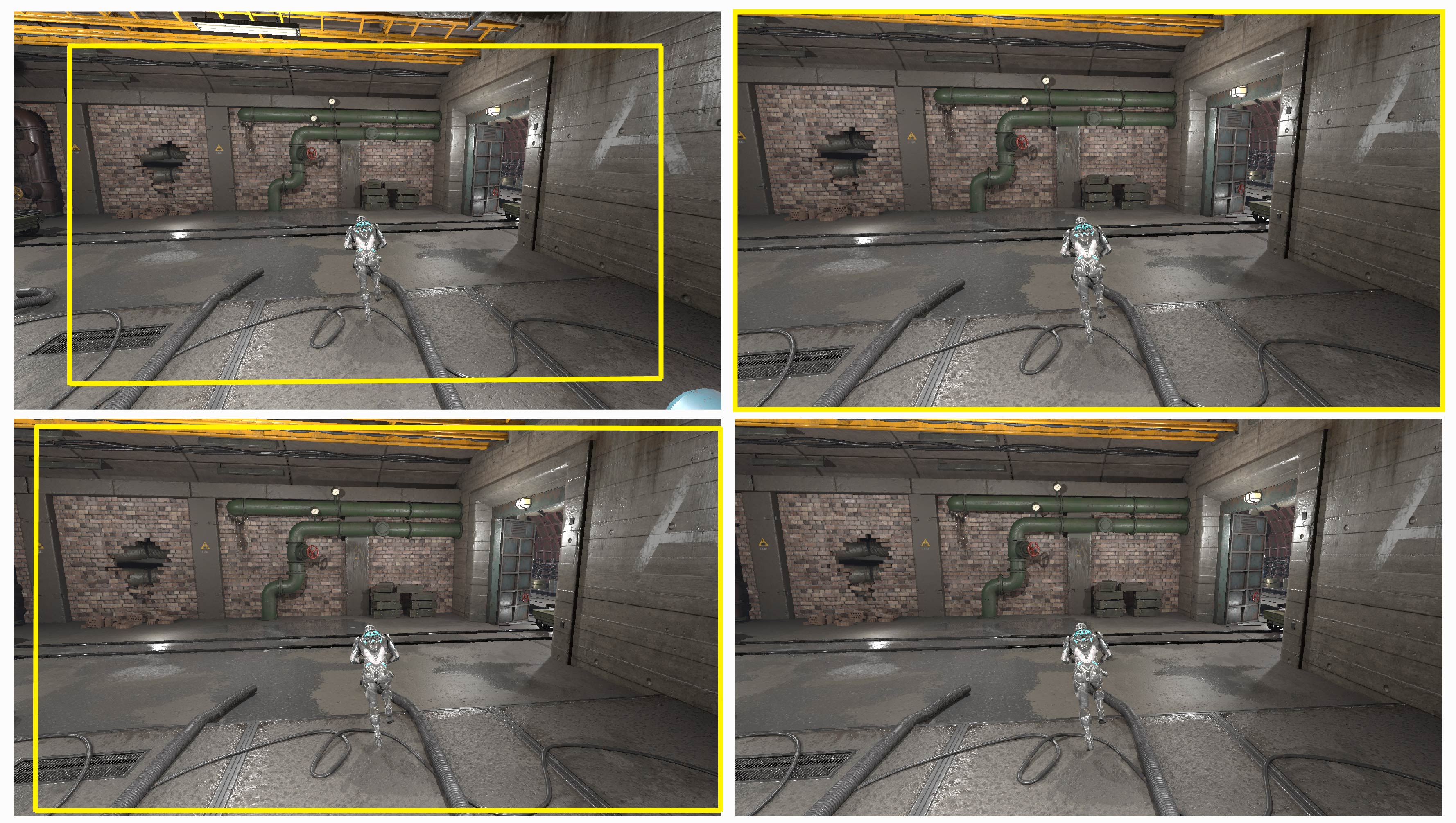}
    \put(36.5, 28.8){\color{white}\tiny{Fixed Enlarged}}
    \put(87, 29.2){\color{white}\tiny{Not Enlarged}}
    \put(33.2, 1.3){\color{white}\tiny{Adaptively Enlarged}}
    \put(88.6, 1.3){\color{white}\tiny{Next View}}
    \end{overpic}
    \caption{Rendered image under different settings. The yellow rectangle shows displayed areas of the frame. All frames are rendered under the same resolution. Our adaptive strategy not only covers the area we need for the next view, but also contains less redundant information.
    }
    \label{fig:dynamicWindow}
\end{figure}
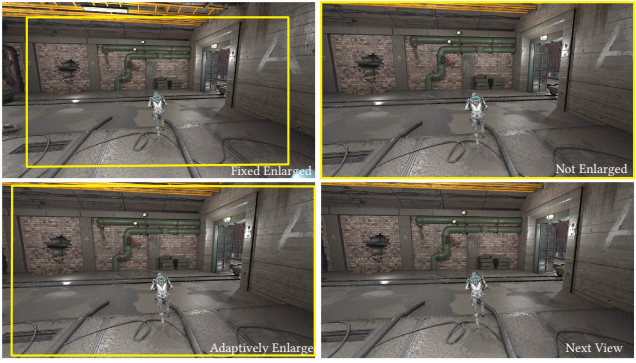

The out-of-screen disocclusion, unlike regions that can be handled by background collection, is never shown in history frames such as continuously rotating camera to the right. These disocclusion areas are on the boundary of the frame and enlarging the original rendering viewport could cover those areas.

A naive way to solve it is to enlarge the field-of-view angle for rendered frames, but many redundant information are included, which leads to blurry results for displayed areas under the same rendering cost.
Instead, we propose an adaptive rendering window strategy to decrease the area of redundant region as shown in Fig.~\ref{fig:dynamicWindow}.

Specifically, when rendering a frame $t$, we estimate the potential areas that will be used for extrapolated frames by two steps: estimate next camera pose and calculate the rendering viewport. Assume camera pose of current frame is $C_t$ and previous rendered frame is $C_{t-1}$, where the pose is formed by three vectors $(v^{\text{pos}}, v^{\text{dir}}, v^{{up}})$. To estimate the camera pose of extrapolated frames $\bar{C}_{t+\alpha}$, we use a similar method as our motion estimation by
\begin{equation}
    \bar{C}_{t+\alpha} = C_t + \alpha \cdot (C_t - C_{t-1}) \\
\end{equation}
where it calculates each vector component separately.

After calculating the estimated camera pose in the next extrapolated frame, the new rendering viewport is approximated based on the union of current camera pose $C_t$ and estimated camera pose $\bar{C}_{t+\alpha}$ rendering areas, which is used for actual rendering. Please refer to appendix for details of calculating actual rendering viewport.

\subsection{Shading Correction Network}
\label{sec:shading_correction}

Previous modules handle the motion of geometries, so we call them geometry aware extrapoltion (GAE) module. However, the motion of shadings is not tracked, and simply ignoring it causes shadings move in low frame rate as analysis in sec.~\ref{sec:challenge}.
Thank to our previous efficient modules which handle geometries motion and disocclusions, we introduce a light neural network called shading correction network (SCN) for non-geometric motion tracking and refinement, which is unlike prior works UPR-Net~\cite{jin2023unified}, IFR-Net~\cite{kong2022ifrnet} and DMVFN~\cite{hu2023dynamic} using large neural networks to estimate the flow for the whole image.

\paragraph{Non-geometric motion detection} To make SCN only focus on the non-geometric motion and shadings, we generate a focus mask to identify the areas that need to be refined and exclude the areas that are already plausible. The focus mask is calculated by the following formula:
\begin{equation}
    \begin{aligned}
    M^{\text{focus}}[x] = & \left(\min_{x'\in N(x)} s(I^{\text{GAE}}[x], I^{\text{gt}}[x']) > 0.5\right) \\
                          &  \land (\hat{M}^{\text{dyn}}[x] = 0) 
    \end{aligned}
\end{equation}
where $s(\cdot, \cdot)$ refers to symmetric mean absolute percentage error (SMAPE), $N(x)$ is the set of $9$ neighborhood pixels and $\hat{M}^{\text{dyn}}$ represents whether a pixel is dynamic. This mask will ignore subtle difference and pixels shifting  to only focus on the shading changes. 

\paragraph{Shading correction network}
\begin{figure}
    \centering
    \begin{overpic}[width=\columnwidth]{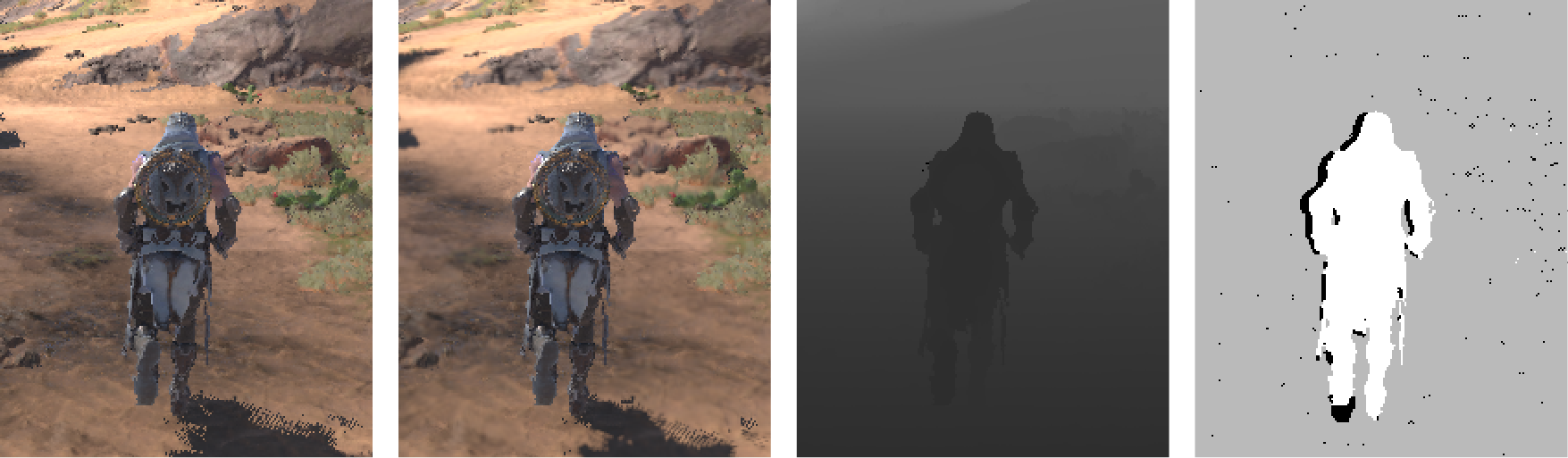}
    \end{overpic}
    \caption{The inputs of shading correction network (SCN). Images from left to right are: the output of GAE module $\bar{I}^{\text{GAE}}_{t+\alpha}$, the backward warped result $I_{t-1\rightarrow t+\alpha}^w$ from frame $t-1$ using motion vectors, the generated depth buffer $\bar{D}_{t+\alpha}$, and the input mask $M_t^{\text{input}}$.}
    \label{fig:scnInput}
\end{figure}
After calculating the focus mask, it guides the shading correction network to only focus on the non-geometric motion. Specifically, the inputs of the network contain: the output of GAE module $\bar{I}_{t+\alpha}^{\text{GAE}}$, the corresponding projected depth buffer $\bar{D}_{t+\alpha}$, the warped frame $I_{t-1\rightarrow t+\alpha}^{w}$ from rendered frame $t-1$ to $t+\alpha$ where ghosting areas are replaced with the correponding areas in $\bar{I}_{t+\alpha}^{\text{GAE}}$, and the input mask $M_{\text{input}}$. An example of the shading correction network's inputs is shown in Fig.~\ref{fig:scnInput}. 

The input mask is generated by our GAE module, where white region indicates dynamic areas, black region indicates disocclusion areas and grey region indicates remaining areas. Warped frame from $t-1$ provides a different shading condition comparing to $\bar{I}_{t+\alpha}^{\text{GAE}}$ in order to calculate the motion of the shading. The remaining invalid areas in $\bar{I}_{t+\alpha}^{\text{GAE}}$ will be filled by down-sampling original image $32$ times before feeding into the network. The final prediction $\bar{I}_{t+\alpha}$ is formulated as
\begin{equation}
    \begin{aligned}
    \bar{I}'_{t+\alpha},\ \bar{M}^{\text{focus}} &= \text{SCN}(\bar{I}_{t+\alpha}^{\text{GAE}}, \bar{D}_{t+\alpha}, I_{t-1\rightarrow t+\alpha}^{w}, M_{t}^{\text{input}}) \\
    \bar{I}_{t+\alpha} &= \bar{I}_{t+\alpha}^{\text{GAE}} \cdot (1 - \bar{M}^{\text{focus}}) + \bar{I}'_{t+\alpha} \cdot \bar{M}_{\text{focus}}
    \end{aligned}
\end{equation}
where $\bar{I}_{t+\alpha}^{\text{GAE}}$ in the second formula is replaced by the ground truth frame $I_{t+\alpha}$ during training. After SCN module, the extrapolated images not only contain correct geometries including dynamic fragments and disocclusions, but also correct shading movement.
Please refer to the appendix for the detailed network architecture, loss functions and training process. 

\section{Experiments}
\label{sec:experiments}

\subsection{Datasets}
\begin{table}
    \begin{center}

        \caption{Scene configuration for training and testing. All frames are captured in 1080p/30fps for inputs and 1080p/60fps for outputs. Our dataset contains less training data and more diverse testing data comparing to previous works~\cite{wu2023extrass, guo2021extranet, wu2023lmv}.}
        \begin{tabular}{c|cccc}
        \toprule

            \multirow{2}{*}{Scenes} & Training & Testing & Training & Testing \\
                                    & Sequences & Sequences & Frames & Frames \\ \hline
                                    Bunker&          2&            1&           2000&           720\\
                                    Park&          2&            1&           2000&           720\\
                                    Future&          2&            1&           2000&           720\\
                                    City&          2&            1&           2000&           720\\
                                    Town&          0&            1&           0&           720\\
                                    Forest&          0&            1&           0&           720\\
                                    Factory&          0&            1&           0&           720\\
                                    Infiltrator&          0&            1&           0&           720\\

        \bottomrule
        \end{tabular}
        \label{tlb:scene}
        
    \end{center}
\end{table}

To demonstrate effectiveness and robustness of our method, we collect 8 difference scenes with different characteristics from Unreal Engine~\cite{unrealengine}. To demonstrate the generalization ability of our method, we use 4 scenes for training and test on all scenes, where 4 scenes are never shown during training. The details of the dataset are shown in Table.~\ref{tlb:scene}. Our collected test scenes are more diverse than previous works~\cite{wu2023extrass, guo2021extranet, wu2023lmv} with few training data in order to show our robustness and generalization ability.

\subsection{Quantitative Metrics}
\label{sec:metrics}

 We evaluate our method with both quantitative metrics and qualitative images/videos to show the comparison. Four metrics are included to show various aspects of our quality: peak signal-to-noise ratio (PSNR), structural similarity index (SSIM), perceptual similarity (LPIPS)~\cite{zhang2018perceptual}, and FovVideoVDP (FvVDP)~\cite{Rafa2021fvvdp}. However, we notice that PNSR and SSIM are less sensitive to blurriness, distortion and temporal flickering since they measure local similarity. LPIPS and FvVDP are more suitable in our case with one measures the whole image perceptual similarity and the other one measures the video perceptual quality. We encourage readers to combine quantitative comparison with image/video qualitative comparison for better understanding.

\subsection{Comparison against Baselines}
To demonstrate the effectiveness of our method, we compare our method with state-of-the-art baselines under three different settings. Note that this is not a fair comparison since frame interpolation and G-buffer dependent extrapolation methods are under easier settings, which means they don't need to handle either disocclusions or motion estimation while our method needs to handle both. Even though, our method still achieves comparable or better results than baselines in general.

UPR-Net~\cite{jin2023unified} and IFR-Net~\cite{kong2022ifrnet} are SOTA video interpolation methods which use optical flow like methods to generate intermediate frames. Offline video interpolation methods~\cite{zhou2022exploring, zhang2023extracting, reda2022film} take more than 100 ms per frame, which is too slow to be used in real-time rendering and irrelevant to our task. DLSS 3.0 and FSR 3 are commercial frameworks where the code and details of implementation are unavailable and it is difficult to obtain the intermediate results for comparison.

ExtraSS~\cite{wu2023extrass} is a joint framework for super resolution and frame extrapolation in real-time rendering with G-buffers for corresponding extrapolated frame. We use ExtraSS-E modules for comparison, which is the extrapolation part of ExtraSS. We choose ExtraSS-E as our baseline instead of LMV~\cite{wu2023lmv} because the latter one requires even more additional G-buffers for rendered and extrapolated frames, which even far away from our goal of G-buffer free frame extrapolation.

DMVFN~\cite{hu2023dynamic} is a video future prediction method which uses current and previous frames to predict the future frame and can be considered as a G-buffer free frame extrapolation baseline.

UPR-Net, IFR-Net and DMVFN are trained on a large scale video dataset and we fine-tune their pre-trained models on our datasets with learning rate $10^{-4}$ for 50 epochs. ExtraSS-E is trained on our datasets from scratch with the same settings as ours.

\subsubsection{Qualitative comparison}

\begin{figure*}
    \centering
    \begin{overpic}[width=\linewidth]{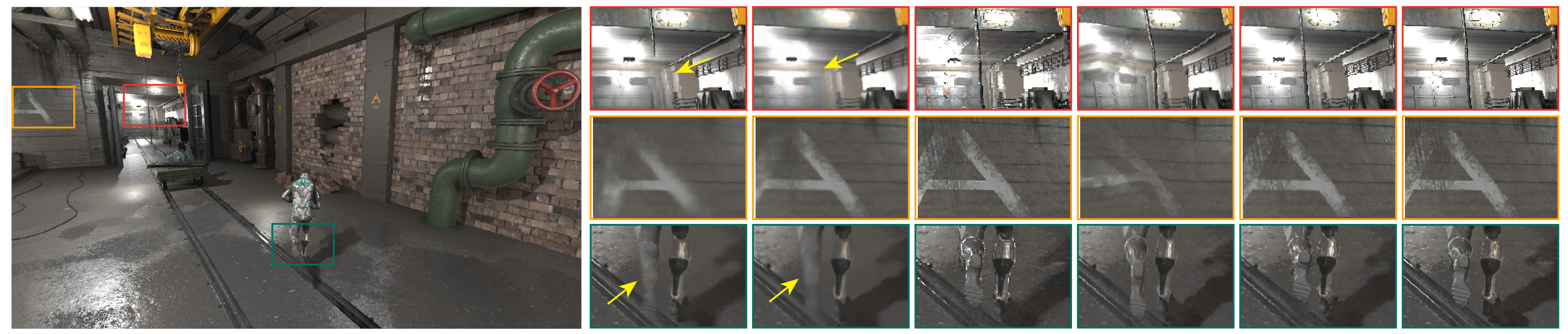}
            \put(14,21.3){\color{black}\small{Ours Full Frame}}
            \put(40,21.3){\color{black}\small{IFR-Net}}
            \put(50.3,21.3){\color{black}\small{UPR-Net}}
            \put(60.4,21.3){\color{black}\small{ExtraSS-E}}
            \put(70.7,21.3){\color{black}\small{DMVFN}}
            \put(82,21.3){\color{black}\small{Ours}}
            \put(93.5,21.3){\color{black}\small{GT}}
            \put(30.1,1){\color{white}\textsc{Bunker}}
    \end{overpic}
    \begin{overpic}[width=\linewidth]{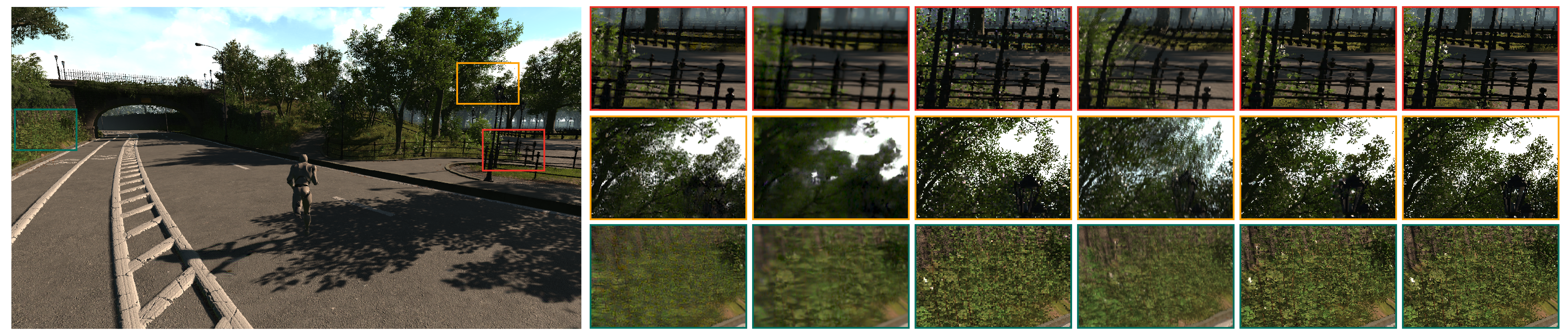}
            \put(32,1){\color{white}\textsc{Park}}
    \end{overpic}    
    \begin{overpic}[width=\linewidth]{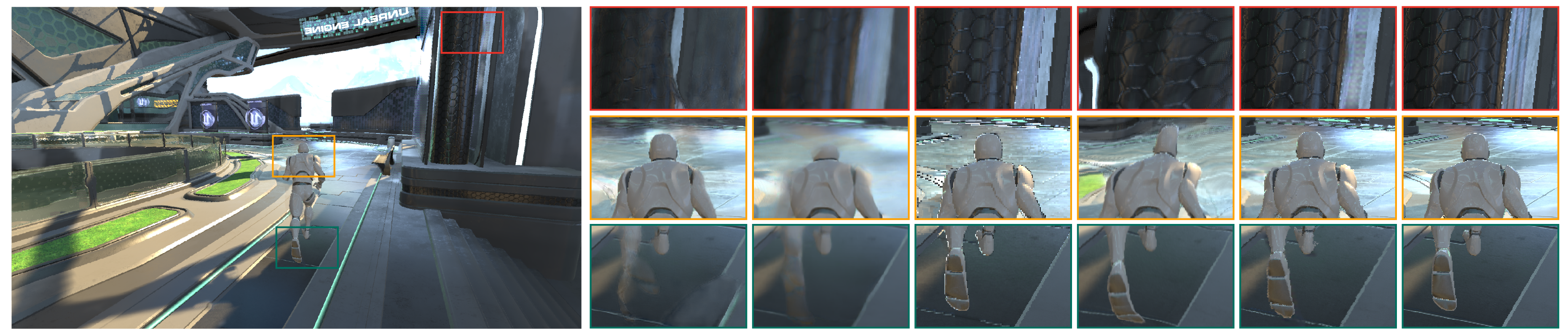}
    \put(31,1){\color{white}\textsc{Future}}
    \end{overpic}
    \begin{overpic}[width=\linewidth]{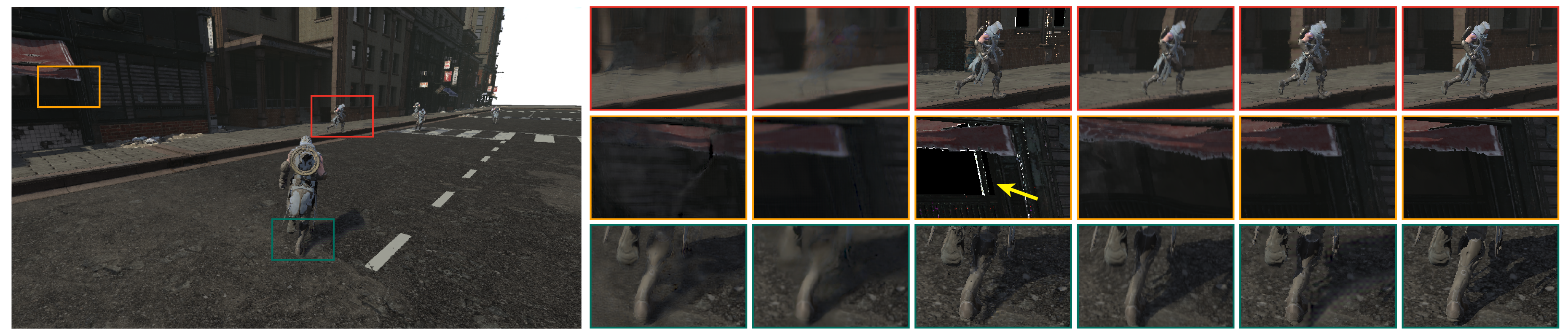}
    \put(32,1){\color{white}\textsc{City}}
    \end{overpic}
    \caption{Qualitative comparison in \textbf{trained} scenes between our method and baseline methods including DMVFN~\cite{hu2023dynamic}, UPR-Net\cite{jin2023unified}, IFR-Net~\cite{kong2022ifrnet} and ExtraSS-E~\cite{wu2023extrass}. DMVFN generates distorted results and cannot generate correct results if the information is missing from two given images. UPR-Net generates over-blurred results and misses thin geometries. ExtraSS-E generates overall good results but fails in translucent materials (windows in the second row). Our method generate detailed extrapolated frames closer to the ground truth with correct geometries and shadings.}
    \label{fig:quality_comparison_train}
\end{figure*}

\begin{figure*}
    \centering
    \begin{overpic}[width=\linewidth]{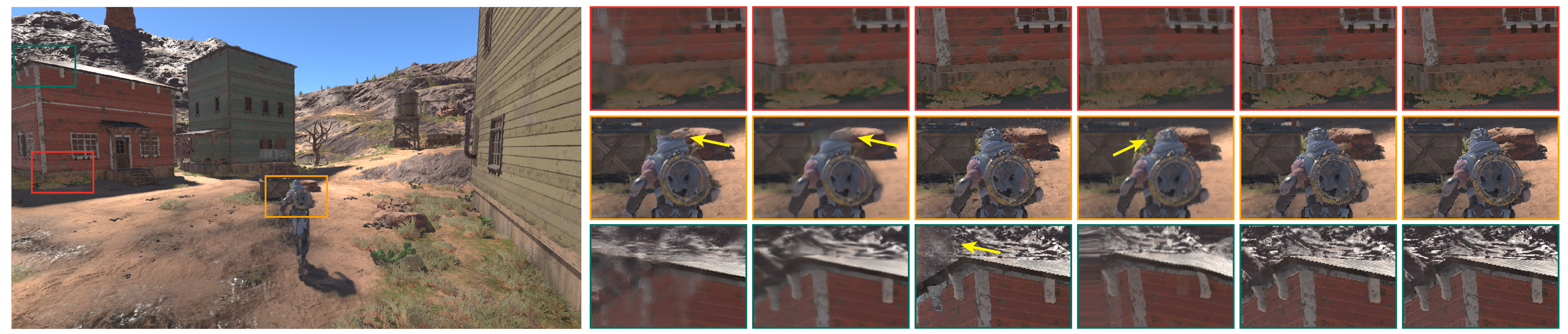}
            \put(14,21.3){\color{black}\small{Ours Full Frame}}
            \put(40,21.3){\color{black}\small{IFR-Net}}
            \put(50.3,21.3){\color{black}\small{UPR-Net}}
            \put(60.4,21.3){\color{black}\small{ExtraSS-E}}
            \put(70.7,21.3){\color{black}\small{DMVFN}}
            \put(82,21.3){\color{black}\small{Ours}}
            \put(93.5,21.3){\color{black}\small{GT}}
            \put(30.1,1){\color{white}\textsc{Town}}
    \end{overpic}
    \begin{overpic}[width=\linewidth]{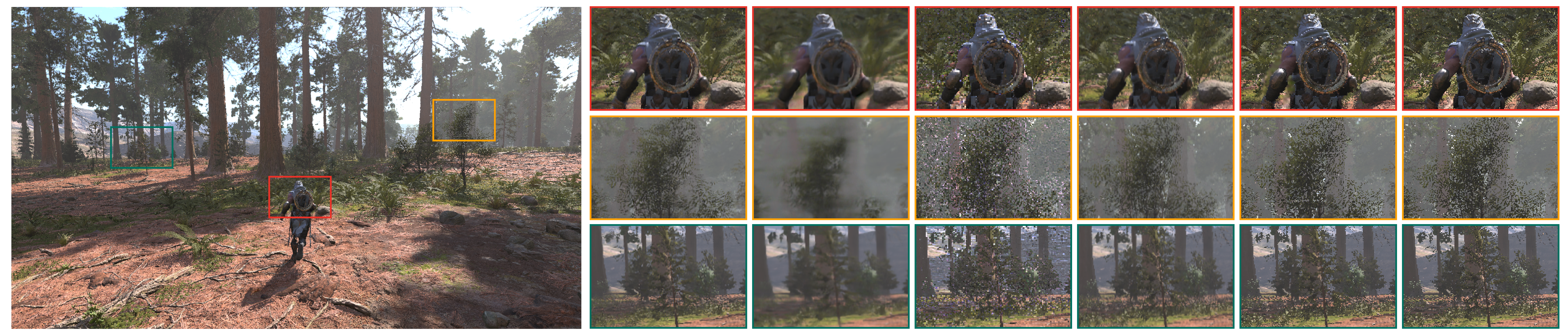}
            \put(30.5,1){\color{white}\textsc{Forest}}
    \end{overpic}    
    \begin{overpic}[width=\linewidth]{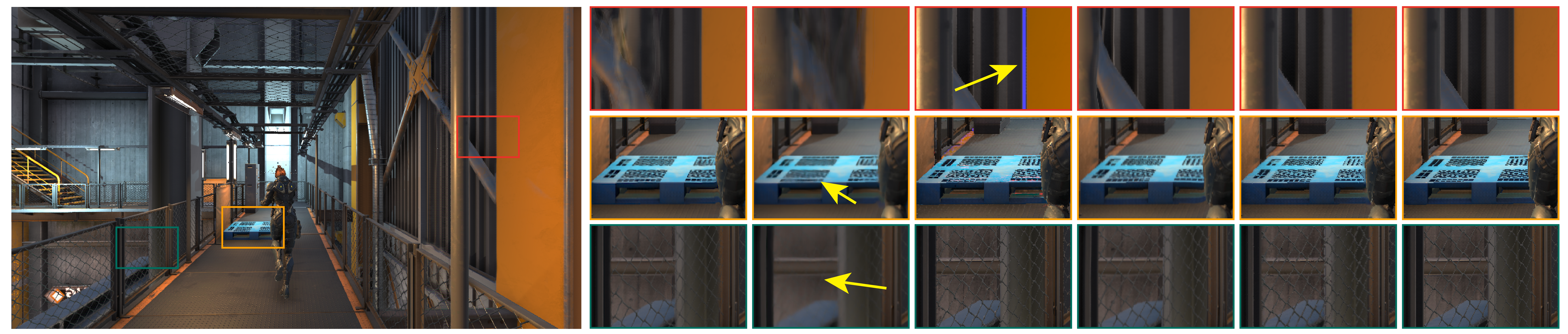}
    \put(29.5,1){\color{white}\textsc{Factory}}
    \end{overpic}
    \begin{overpic}[width=\linewidth]{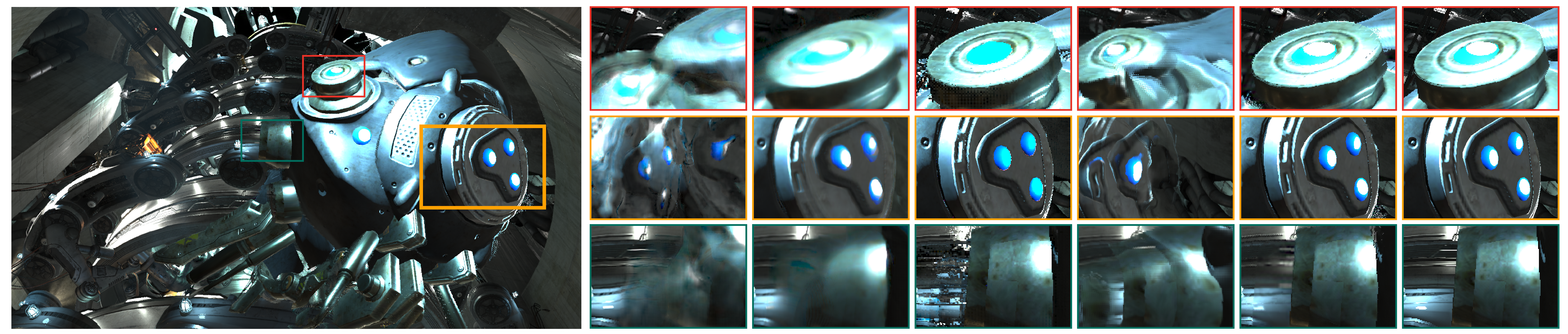}
    \put(27,1){\color{white}\textsc{Infiltrator}}
    \end{overpic}
    \caption{Qualitative comparison in \textbf{test} scenes between our method and baseline methods including DMVFN~\cite{hu2023dynamic}, UPR-Net\cite{jin2023unified}, IFR-Net~\cite{kong2022ifrnet} and ExtraSS-E~\cite{wu2023extrass}. Our method still shows comparable or better visual quality with less distortion, blurriness and artifacts even though there are some big gap between training scenes and the test scenes.}
    \label{fig:quality_comparison_test}
\end{figure*}

The qualitative comparison is shown in Fig.~\ref{fig:quality_comparison_train} (trained scenes) and Fig.~\ref{fig:quality_comparison_test} (not trained scenes). 
DMVFN generates highly distorted results when motion is large and can not generate correct results for the areas that don't have corresponding information in prior two frames. 
Frame interpolation methods UPR-Net and IFR-Net cannot track the motion of thin geometries, so thin geometries are usually missing in this case. Besides, their estimated optical flows are not accurate enough to generate clear results so their results are usually over-blurred or even severely distorted in some cases.
ExtraSS-E uses ground truth G-buffers to guide the generation of extrapolated frames, which is usually more stable and contains more details. However, it fails in translucent materials (windows in \textsc{City}) and generates flickering results without its own super sampling module. Our method generates more stable frames with sharper details and less distortions.

\begin{table*}
    \centering
    
    \setlength{\tabcolsep}{2.5mm}{
    \renewcommand{\arraystretch}{1.05} {
\caption{Quantitative comparison with UPR-Net~\cite{jin2023unified}, IFR-Net~\cite{kong2022ifrnet}, DMVFN~\cite{hu2023dynamic} and ExtraSS-E~\cite{wu2023extrass} under PSNR, SSIM, LPIPS, and FvVDP. Our method shows comparable quality with interpolation methods UPR-Net and IFR-Net, and G-buffer dependent extrapolation method ExtraSS-E under PSNR and SSIM metrics. Besides, our method shows better perceptual quality than interpolation methods and G-buffer dependent method since these baselines are over-blurred, distorted or flickering as anaylzed in qualitative comparison. Our method also outperforms G-buffer free extrapolation baseline DMVFN in all aspects. The values of SSIM and LPIPS are scaled by $10^2$. }
\label{tlb:quantitative_comparison}
    \begin{tabular}{ccc|cccc|cccc|c}
        \toprule
         \multicolumn{3}{c|}{}  &\multicolumn{4}{c|}{Trained}&\multicolumn{4}{c|}{Not trained} &\\
         & Type & Method  &  Bunker&  Park&  Future&  City&  Town&  Forest&  Factory& Infiltrator & Average\\
         \hline
         \multirow{5}{1em}{\rotatebox[origin=c]{90}{PSNR $\uparrow$}} & \multirow{2}{2.5em}{Interp.} & IFR &   25.55&  18.37&  28.66&  26.16&  27.41&  19.62&  22.40&  25.78&24.24\\
                                                           &                              & UPR&   26.35&  18.52&  \textbf{28.75}&  27.99&  \textbf{27.75}&  19.60&  23.87&  \textbf{26.12}&\textbf{24.87}\\
                                    \cline{2-12}
                                                           &       G-buf Extrap.                    & ExSS-E&   26.44&  \textbf{24.75}&  25.77&  24.38&  27.45&  \textbf{21.71}&  23.10&  23.53&24.64\\
                                    \cline{2-12}
                                                           & \multirow{2}{2.5em}{Extrap.} & DMVFN &   23.79&  16.89&  23.69&  24.77&  24.44&  17.00&  20.38&  22.84&21.73\\
                                                           &                              & Ours &   \textbf{27.84}&  17.04&  26.33&  \textbf{29.77}&  26.09&  18.21&  \textbf{24.50}&  23.78&24.20\\ 
                                    \midrule
         \multirow{5}{1em}{\rotatebox[origin=c]{90}{SSIM $\uparrow$}}& \multirow{2}{2.5em}{Interp.} & IFR &   86.78&  71.43&  \textbf{94.82}&  84.84&  91.16&  70.23&  84.66&  91.21&84.39\\
                                                           &                              & UPR&   88.21&  72.44&  94.65&  88.26&  90.58&  67.55&  86.16&  \textbf{91.77}&84.95\\
                                    \cline{2-12}
                                                           &       G-buf Extrap.           & ExSS-E&   91.17&  \textbf{86.79}&  91.68&  88.92&  \textbf{92.32}&  \textbf{78.57}&  88.45&  85.51&\textbf{87.93}\\
                                    \cline{2-12}
                                                           & \multirow{2}{2.5em}{Extrap.} & DMVFN &   82.39&  61.18&  87.46&  81.43&  81.63&  51.97&  76.25&  86.39&76.09\\
                                                           &                              & Ours &   \textbf{93.50}&  73.80&  93.46&  \textbf{93.49}&  89.04&  65.25&  \textbf{89.55}&  89.37&85.93\\ 
                                    \midrule
         \multirow{5}{1em}{\rotatebox[origin=c]{90}{LPIPS $\downarrow$}} & \multirow{2}{2.5em}{Interp.} & IFR &   15.62&  24.19&  10.67&  22.86&  11.89&  28.00&  17.95&  12.88&18.01\\
                                                           &                              & UPR&   22.49&  42.44&  17.15&  24.05&  22.99&  52.81&  26.15&  19.19&28.41\\
                                    \cline{2-12}
                                                           &       G-buf Extrap.            & ExSS-E&   12.22&  14.63&  14.73&  17.38&  \textbf{7.79}&  \textbf{17.78}&  15.72&  24.26&15.56\\
                                    \cline{2-12}
                                                           & \multirow{2}{2.5em}{Extrap.} & DMVFN &   16.89&  28.12&  12.83&  20.06&  16.65&  35.44&  20.79&  15.42&20.78\\
                                                           &                              & Ours &   \textbf{6.68}&  \textbf{14.02}&  \textbf{5.74}&  \textbf{7.24}&  7.98&  16.98&  \textbf{9.88}&  \textbf{8.81}&\textbf{9.67}\\ 
                                    \midrule
         \multirow{5}{1em}{\rotatebox[origin=c]{90}{FvVDP $\uparrow$}} & \multirow{2}{2.5em}{Interp.} & IFR &   7.98&  6.77&  5.36&  7.20&  8.35&  7.07&  6.84&  7.77&7.17\\
                                                           &                              & UPR&   8.52&  6.85&  \textbf{5.40}&  8.45&  \textbf{8.75}&  7.16&  7.24&  8.05&7.55\\
                                    \cline{2-12}
                                                           &       G-buf Extrap.           & ExSS-E&   8.21&  \textbf{7.40}&  5.30&  5.96&  8.11&  \textbf{7.38}&  7.39&  7.78&7.19\\
                                    \cline{2-12}
                                                           & \multirow{2}{2.5em}{Extrap.} & DMVFN &   7.25&  6.57&  5.37&  6.85&  7.88&  6.43&  6.49&  7.56&6.80\\
                                                           &                              & Ours &   \textbf{8.65}&  6.97&  5.37&  \textbf{8.58}&  8.65&  7.05&  \textbf{7.87}&  \textbf{8.09}&\textbf{7.65}\\ 
                                    \hline
        \bottomrule
    \end{tabular}
    }
    }
\end{table*}

\subsubsection{Quantitative comparison}
Table.~\ref{tlb:quantitative_comparison} shows the quantitative comparison against baselines. As discussed in sec.~\ref{sec:metrics}, PSNR and SSIM metrics are not always reliable to evaluate the quality of the generated frames. UPR-Net and IFR-Net shows severe distortion and missing geometries as shown in previous qualitative comparison, although they have higher PSNR and SSIM metrics in some scenes.  ExtraSS-E shows marginaly better results in scenes with complex geometries such as \textsc{Forest} and \textsc{Park} since G-buffers provide strong clues for the generation of extrapolated frames but the time of generating G-buffer in those scenes is usually long. Besides, it fails with scenes with translucent materials like \textsc{City} and \textsc{Future}. DMVFN shows significant lower PSNR and SSIM metrics than other methods since it struggles in disocclusion areas and generates severelt distorted results.

Besides PSNR and SSIM metrics, our method shows better results in LPIPS and FvVDP metrics which are more perceptual and suitable for evaluating the quality of the generated frames, and more consistent with qualitative image and video comparison. This is because our method generates more stable and plausible frames than other baselines, which is important in real-time rendering applications since people notice the flickering and distortion more than pixel level differences.

Based on our quantitative and qualitative evaluation, GFFE is significantly better than G-buffer free baselines in all aspects, and shows comparable results with G-buffer dependent and interpolation methods with better perceptual quality in LPIPS and FvVDP metrics. 

\subsection{Generalization}
Our framework is trained on 4 scenes and tested on 8 scenes, where 4 scenes are never shown during training. Despite the limited dataset, our method generates stable and plausible results in all scenes. This is because our hybrid modules are robust to different scenes and can handle disocclusions and motion estimation well, instead of using a single neural network to handle all problems which requires a large scale dataset for training. Therefore, we consider our method has good generalization ability and robustness to various different scenes.

\subsection{Performance}
\label{sec:performance}
\begin{table}
    \begin{center}

    \setlength{\tabcolsep}{2.5mm}{
    \renewcommand{\arraystretch}{1.3} {
        \caption{Runtime (ms) for all methods to generate 1080p frames, $+$ means not including the time of generating G-buffers, see Table.~\ref{tlb:performance_gbuf}.}
        \begin{tabular}{ c c c c c }
        \toprule[1pt]
                                             
            UPRNet & IFRNet & DMVFN & ExSS-E  & Ours-Full \\ \hline
            43.04 & 19.50 & 20.57 & 4.18+  & 
            \textbf{6.62} \\

        \bottomrule[1pt]
        \end{tabular}
        \label{tlb:performance}
    }
        
    }
    \end{center}
\end{table}

\begin{table}
    \begin{center}

    \setlength{\tabcolsep}{3.5mm}{
    \renewcommand{\arraystretch}{1.2} {
        \caption{Runtime (ms) breakdown for our framework under different resolutions. Misc mainly includes adjusting display window and maintaining correct motion vectors.}
        \begin{tabular}{ c | c c c }
        \toprule
                                             
                 & 540p & 720p & 1080p \\ \hline
            BG Collection & 0.34& 0.54& 1.13    \\
            History Track & 0.27& 0.49& 1.04  \\
            Misc          & 0.09& 0.14& 0.37  \\ \hline
            BG Projection & 0.16& 0.32& 0.76 \\
            Position Pred. & 0.07& 0.11& 0.22 \\ 
            Warp  & 0.51& 0.76& 0.80 \\ \hline
            SCN         & 0.90& 1.30& 2.30 \\ \hline
            Total       & 2.34& 3.66& 6.62 \\

        \bottomrule
        \end{tabular}
        \label{tlb:performance_ours}
    }
        
    }
    \end{center}
\end{table}

\begin{table}
    \begin{center}

    \setlength{\tabcolsep}{3.5mm}{
    \renewcommand{\arraystretch}{1.0} {
        \caption{G-buffer generation time (ms) under 1080p for different scenes. For non-experimented scenes in products, the time may even exceed $10$ ms.}
        \begin{tabular}{ c | c c c c }
        \toprule[1pt]
                & Bunker & Park & Future & City \\
            Time &   0.35     &  8.23    &   2.85    &   0.40    \\
            \midrule 
                & Town    & Forest & Factory & Infiltrator \\
            Time  &   2.02     &  2.61    &  1.91     &   0.96    \\

        \bottomrule[1pt]
        \end{tabular}
        \label{tlb:performance_gbuf}
    }
        
    }
    \end{center}
\end{table}

Performance is an important factor for real-time rendering applications. We used a machine with NVIDIA RTX 4070Ti Super GPU and Ryzen 9 5900X CPU for inference. The non-neural modules (GAE) of our method are implemented in NVIDIA Falcor~\cite{Kallweit:2022:falcor} renderer. All neural networks including baselines are trained under PyTorch framework and converted into TensorRT~\cite{tensorrt} with FP16 precision for inference.  

Table.~\ref{tlb:performance_ours} shows the break down run time of our method under different resolutions. Note that our method is designed to be applied in the post-processing stage and the performance is not affected by complexity of the scene.

Table.~\ref{tlb:performance} shows the run times for all methods. Previous frame interpolation methods UPR-Net and IFR-Net and extrapolation method DMVFN are much slower than our method, since neural networks are usually slow comparing to heuristic methods. 
ExtraSS-E is faster than our method since it uses G-buffers to guide the generation of extrapolated frames. However, the time of generating G-buffers is not included in the runtime of ExtraSS-E, which varies depending on the complexity of the scene. Table.~\ref{tlb:performance_gbuf} shows the time of generating G-buffers for different scenes, where complex geometries scene like \textsc{Forest} and \textsc{Park} takes longer time to generate G-buffers. Note that the scenes that ExtraSS-E is better than ours are usually the scenes with complex geometries. 

Although there are some dedicated hardware or software optimizations could be applied to accelerate the runtime performance, all methods are tested under the same environment and settings without dedicated optimizations, so any optimizations applied to baselines can also be applied to our method to achieve better performance.

\section{Ablation Study}
\label{sec:ablation}

\begin{table}
    \begin{center}

    \setlength{\tabcolsep}{2.5mm}{
    \renewcommand{\arraystretch}{1.1} {
        \caption{Ablation study on our designed modules. Numbers are averaged for all scenes. SSIM and LPIPS numbers are scaled by $10^2$. ME = Motion Estimation, BGC = Hierarchical Background Collection, AW = Adaptive Render Window, SCN = Shading Correction Network, FM = Focus Mask}
        \begin{tabular}{c|cccc}
        \toprule

            & PSNR$\uparrow$ & SSIM$\uparrow$ & LPIPS$\downarrow$ & FvVDP$\uparrow$ \\ \hline
        w/o ME & 23.44& 85.21& 10.36& 7.57\\
        w/o BGC & 24.11 & 85.91 & 9.79 & 7.62\\
        w/o AW & 24.15& 85.91& 9.71& 7.62\\
        w/o SCN & 23.93 & 85.86 & \textbf{9.23} & 7.60\\
        w/o FM  & 23.86  & 81.39&      35.63&     7.49\\
        Ours Full & \textbf{24.20} & \textbf{85.93} & 9.67& \textbf{7.65} \\

        \bottomrule
        \end{tabular}
        \label{tlb:ablation}
    }
        
    }
    \end{center}
\end{table}

Our framework is a complete pipeline that consists of several modules for G-buffer free frame extrapolation. In this section, we analyze the effectiveness and importance of each module in our framework to demonstrate the necessity of each module. Table.~\ref{tlb:ablation} show the quantitative metrics of removing our designed modules, and more qualititative results will included in following sections and the supplementary video.

\subsection{Motion Estimation}
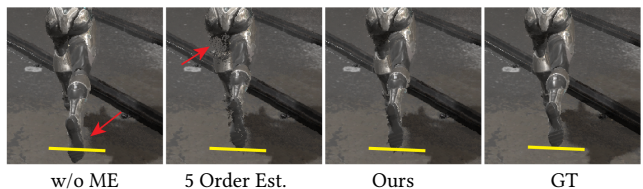
\begin{figure}
    \centering
    \begin{overpic}[width=\columnwidth]{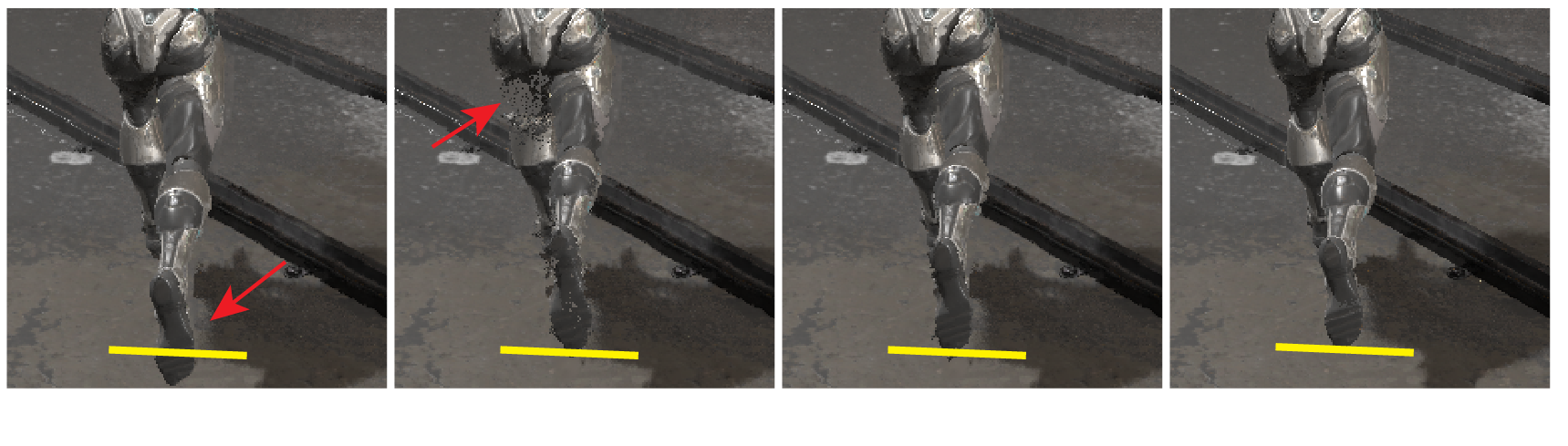}
    \put(7.2, 0){\color{black}\small{w/o ME}}
    \put(28, 0){\color{black}\small{$5$ Order Est.}}
    \put(57.2, 0){\color{black}\small{Ours}}
    \put(85, 0){\color{black}\small{GT}}
    \end{overpic}
    \caption{Ablation study of the motion estimation (ME) module. Without this module, the geometry will move. With higher order polynomials to estimate the motion, it diverges and is unstable. Our approach estimates the motion for dynamic objects more plausibly. }
    \label{fig:abl_me}
\end{figure}
Motion estimation tracks the motion of dynamic fragments and project to extrapolated frames. Linear motion in the world space is used for estimate the next world position in the extrapolated frames. Higher order polynomials could be used but with worse quality in our experiments. As shown in Fig.~\ref{fig:abl_me}, dynamic objects are not moving without motion estimation module. Higher order polynomials estimation diverges and generates artifacts. Our module efficiently generates plausible motions for dynamic objects.

\subsection{Background Collection}
\begin{figure}
    \centering
    \begin{overpic}[width=\columnwidth]{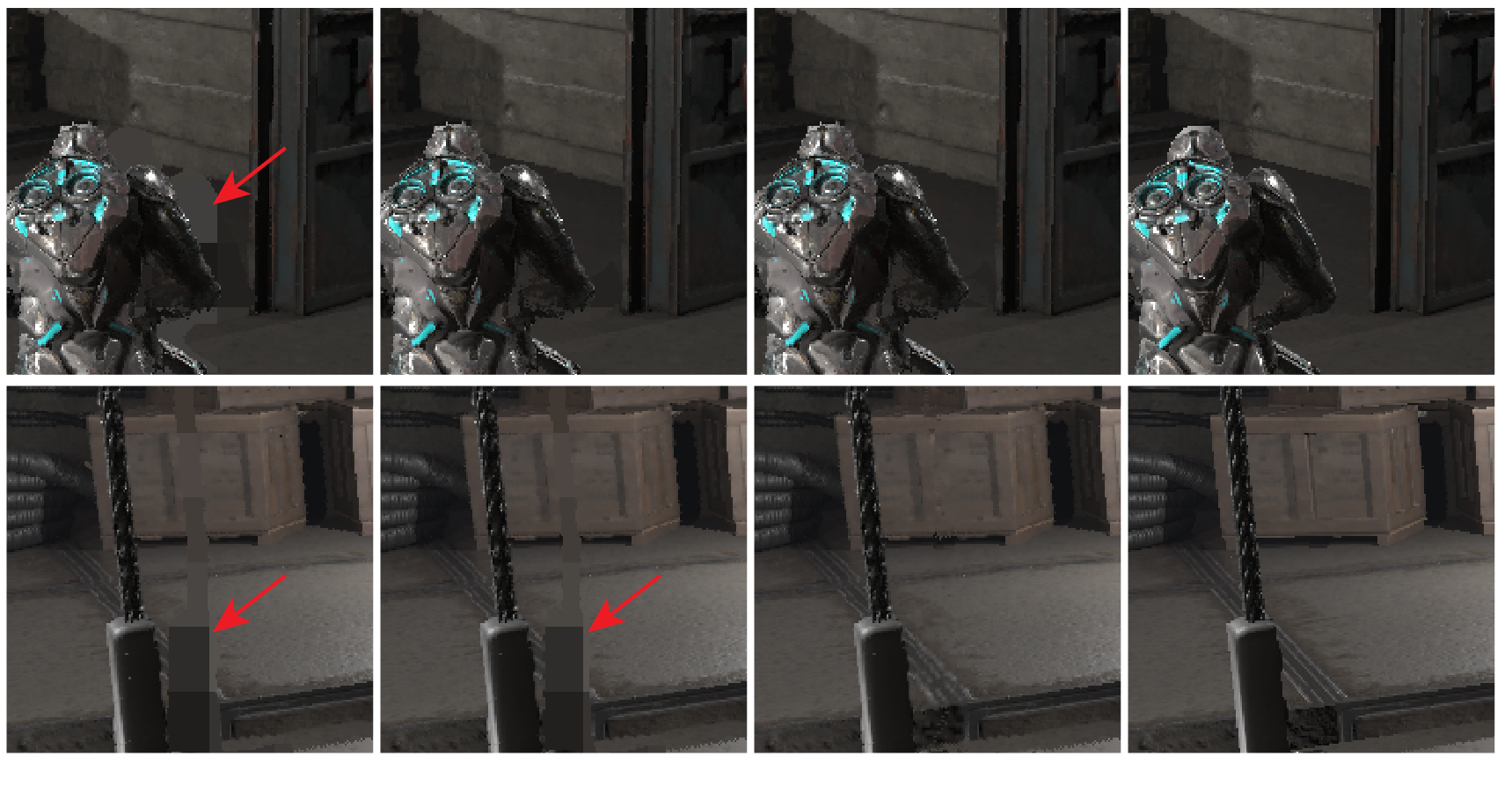}
    \put(7.2, 0){\color{black}\small{w/o BC}}
    \put(33, 0){\color{black}\small{1 Layer}}
    \put(56.8, 0){\color{black}\small{2 Layers}}
    \put(85, 0){\color{black}\small{GT}}
    \end{overpic}
    \caption{Ablation study of the hierarchical background collection module. Without background collection, the disocclusion areas are not handled at all. With one layer background, it only captures background behind dynamic objects. With our two-layers background collection, it not only recovers disocclusion behind dynamic objects, but also static disocclusions behind static objects. }
    \label{fig:abl_bgc}
\end{figure}
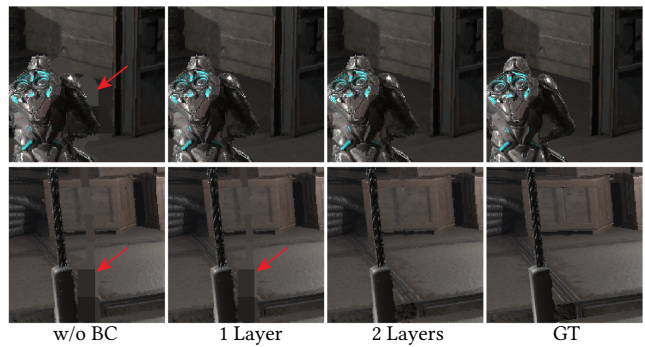

Background collection addresses static disocclusions and dynamic disocclusions. Without such module and directly to guess what is in the disocclusion areas, the results are usually in lower quality. Our hierarchical background collection module collects multiple layers background to handle different levels disocclusions. Fig.~\ref{fig:abl_bgc} shows the comparison between results with and without background collection. We can see that our complete background fixes not only disocclusions behind dynamic objects but also for the disocclusions behind the static objects due to camera motion. 

\subsection{Adaptive Rendering Window}
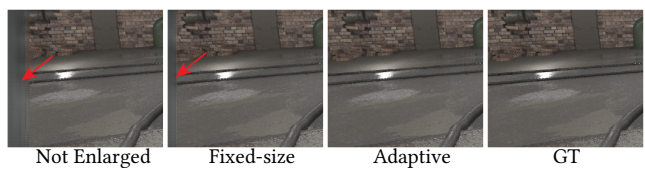
\begin{figure}
    \centering
    \begin{overpic}[width=\columnwidth]{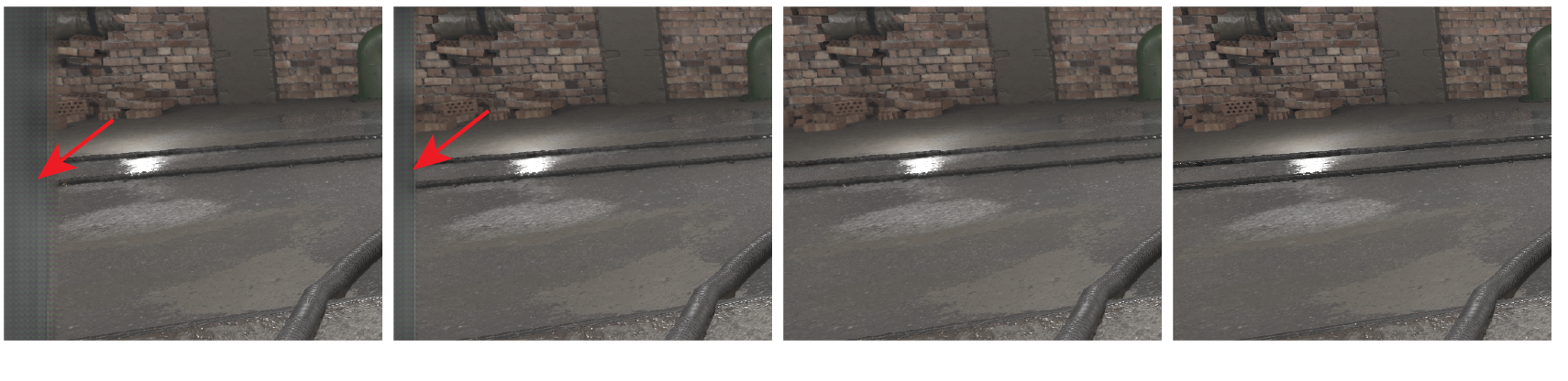}
    \put(4.6, 0){\color{black}\small{Not Enlarged}}
    \put(31.5, 0){\color{black}\small{Fixed-size}}
    \put(57, 0){\color{black}\small{Adaptive}}
    \put(85, 0){\color{black}\small{GT}}
    \end{overpic}
    \caption{Ablation study of adaptive rendering windows. Invalid region appears at the boundary of the image without the adaptive windows. Fixed-size enlarged rendering window contains more redundant and less useful information for extrapolation. Our adaptive strategy can adjust rendering window dynamically for better extrapolation.}
    \label{fig:abl_adaptive}
\end{figure}
We compare our adaptive strategy with fixed enlarged window and not enlarged window. Fig.~\ref{fig:abl_adaptive} shows the comparison between these three methods, and our method covers more disocclusions since our rendering window is adaptively adjusted based on camera motion. 
 
\subsection{Shading Correction Network}
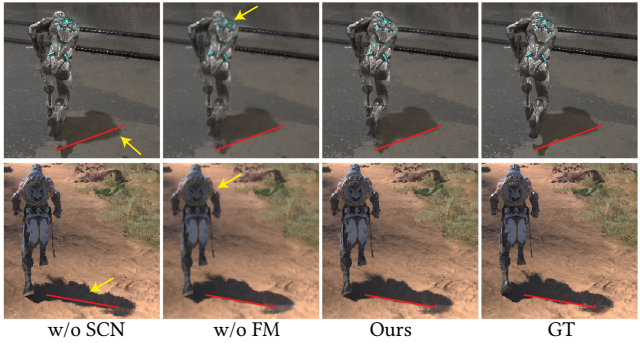
\begin{figure}
    \centering
    \begin{overpic}[width=\columnwidth]{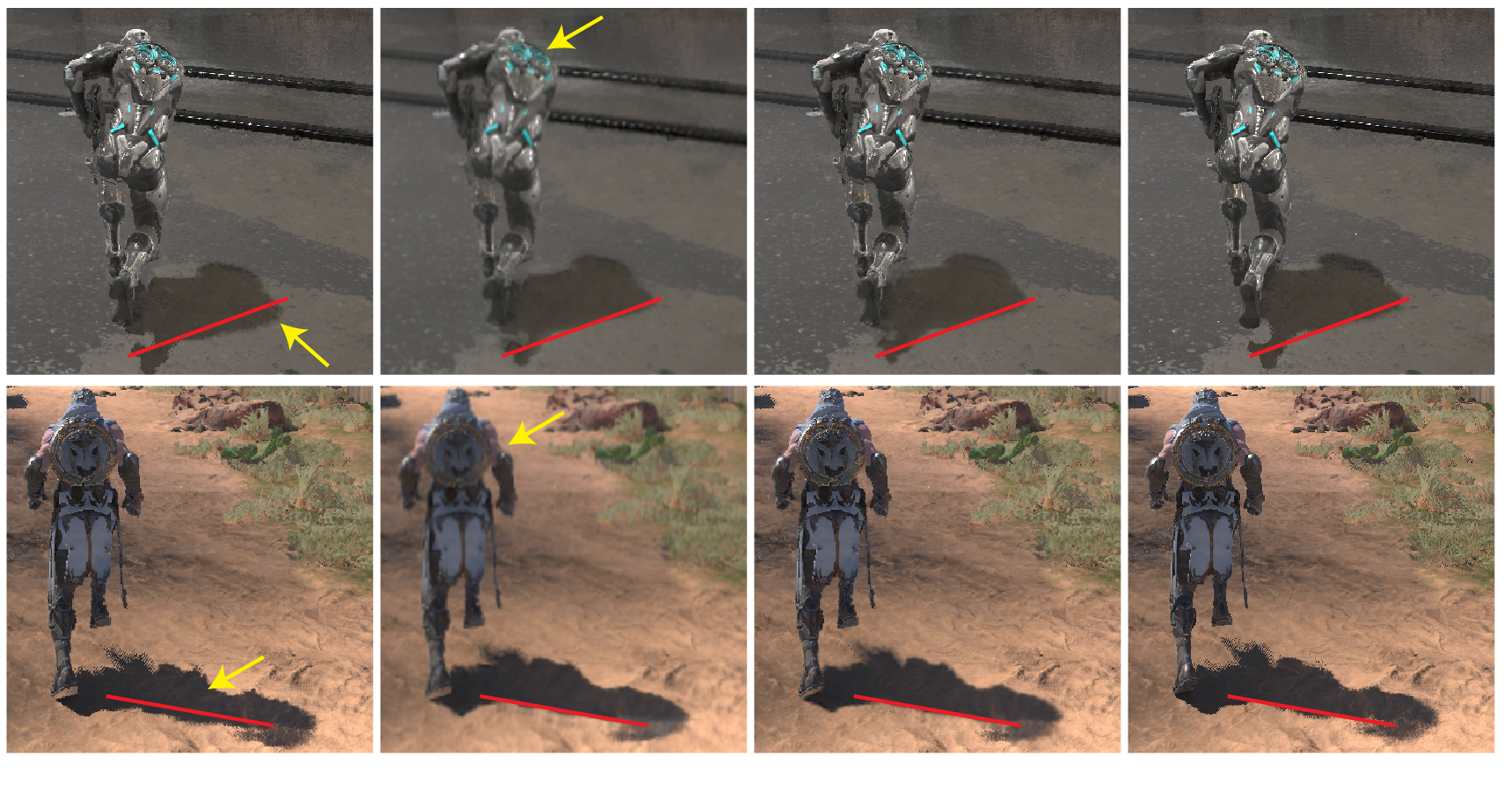}
    \put(7.2, 0){\color{black}\small{w/o SCN}}
    \put(33, 0){\color{black}\small{w/o FM}}
    \put(57.3, 0){\color{black}\small{Ours}}
    \put(85, 0){\color{black}\small{GT}}
    \end{overpic}
    \caption{Ablation study of the shading correction network (SCN). FM = Focus Mask. Without SCN, non-geometric motions are not tracked, so shadows are not moving in extrapolated frame. Without the focus mask, directly predicting the final refined frame will blurs the whole image. Our SCN with focus mask not only fix non-geometric motion, but also keeps sharp details. }
    \label{fig:abl_scn}
\end{figure}

Our SCN module mainly fixes the lagging issue of non-geometries motion including shadows and reflections. As discuss in previous work~\cite{guo2021extranet}, although such effects have small impact in metrics or even slightly worse (LPIPS), it is noticeable in human perception and important for high quality rendering. 

Fig.~\ref{fig:abl_scn} shows the comparison between not using SCN, without focus mask, and with our full SCN module. Without SCN, the shadows and reflections are not moving due to missing motion, leading low frame rate feeling in those areas. Without focus mask, the neural network tries to refine the whole image, which blurs the overall details. Our full SCN module can detect the areas that need to be refined and only refine those areas and do not blur other areas. For better visualization and comparison of this ablation study, please refer to the supplementary video to see how it affects the final results for continuous frames.

\section{Discussion}

\subsection{Anti-aliasing and Super Resolution}
\label{sec:dlss}

\begin{figure}
    \centering
    \begin{overpic}[width=\columnwidth]{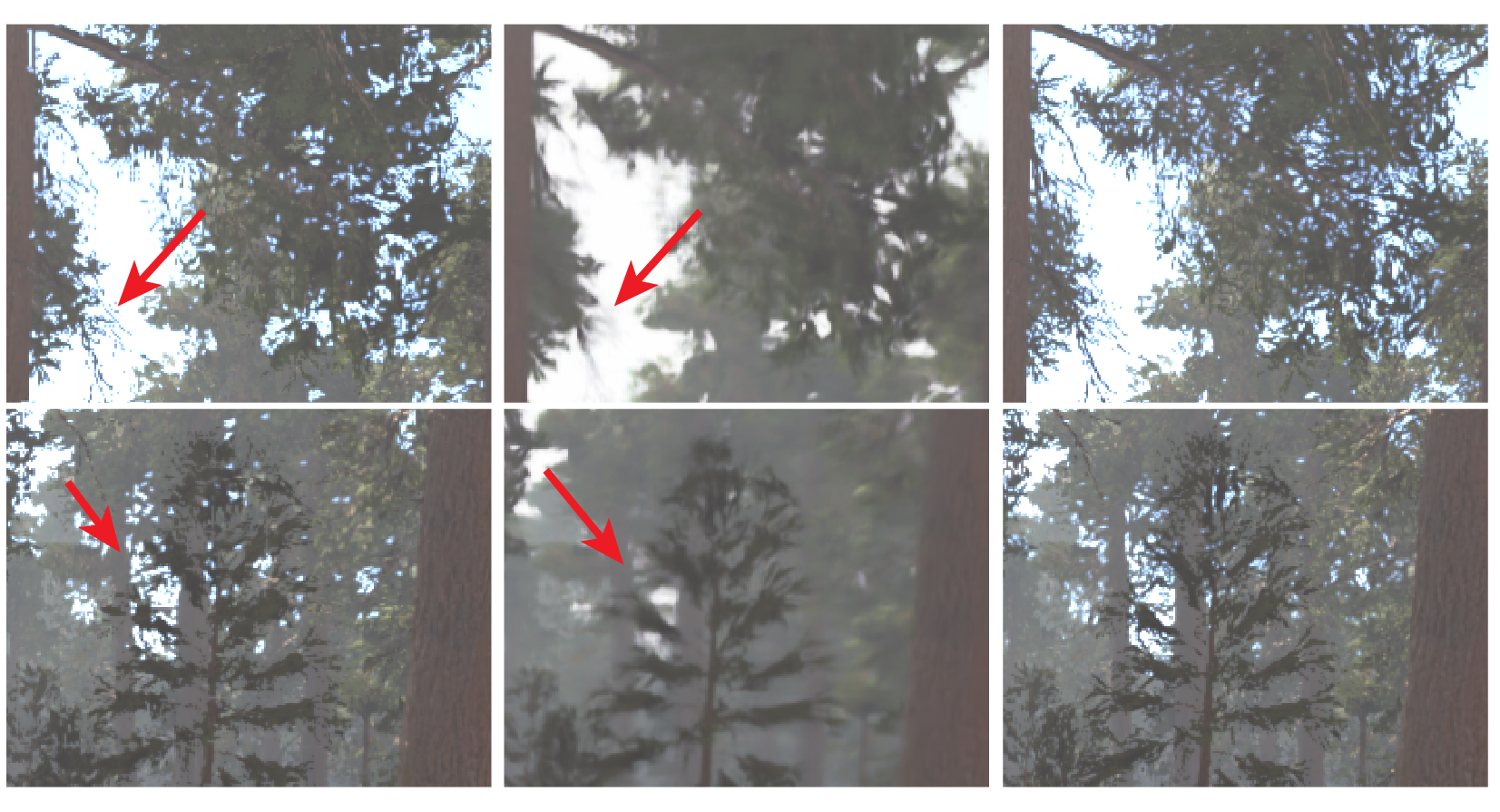}
    \put(8, 53){\color{black}\small{Ours-DLSS}}
    \put(45, 53){\color{black}\small{UPR-DLSS}}
    \put(78, 53){\color{black}\small{GT-DLSS}}
    \end{overpic}
    \caption{Results of integrating DLSS 2 with ours and UPR~\cite{jin2023unified}. Our results contain more details while UPR with DLSS 2 still generates blurred results.}
    \label{fig:dlss}
\end{figure}

Our framework, unlike UPR-Net, IFR-Net and DMVFN, generates not only extrapolated shaded frames, but also the corresponding depth buffer and motion vectors between the extrapolated frames and rendered frames. This indicates that the generated frames can be considered the same as other rendered frames to apply additional anti-aliasing or super resolution techniques. 

Super sampling techniques, including DLSS\cite{liu2020dlss}, XeSS\cite{xess}, FSR\cite{fsr}, have shown high quality results in generating higher resolution frames from lower resolution frames efficiently which are widely used in real-time rendering to improve the visual quality. Our method with generated depth and motion vectors can be easily integrated with such super resolution techniques to generate higher quality frames. Fig.~\ref{fig:dlss} shows the comparison between our method and UPR~\cite{jin2023unified} of using DLSS on \textsc{Forest} with complex geometries. Our results contains more details and UPR tends to over-blur them. Ground truth depth and motion vectors are used for baselines.

\subsection{Practical Choice}
We show breakdown performance and ablation studies in previous section to demonstrate the usage to each module. Each module in our framework is relatively independent and can be removed or replaced by better modules in future if needed. For example, for low end devices such as mobiles, neural network module SCN could be removed since the shading changes are usually simpler, so the integration is easier and performance is better with some degradation in quality. Our framework is flexible to be adjusted in various applications based on needs.

\subsection{Limitations}
\label{sec:limitation}

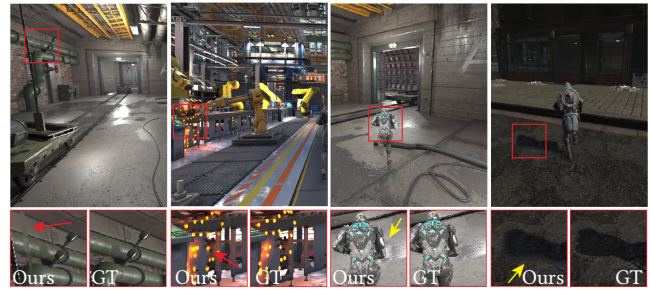
\begin{figure}
    \centering
    \begin{overpic}[width=\columnwidth]{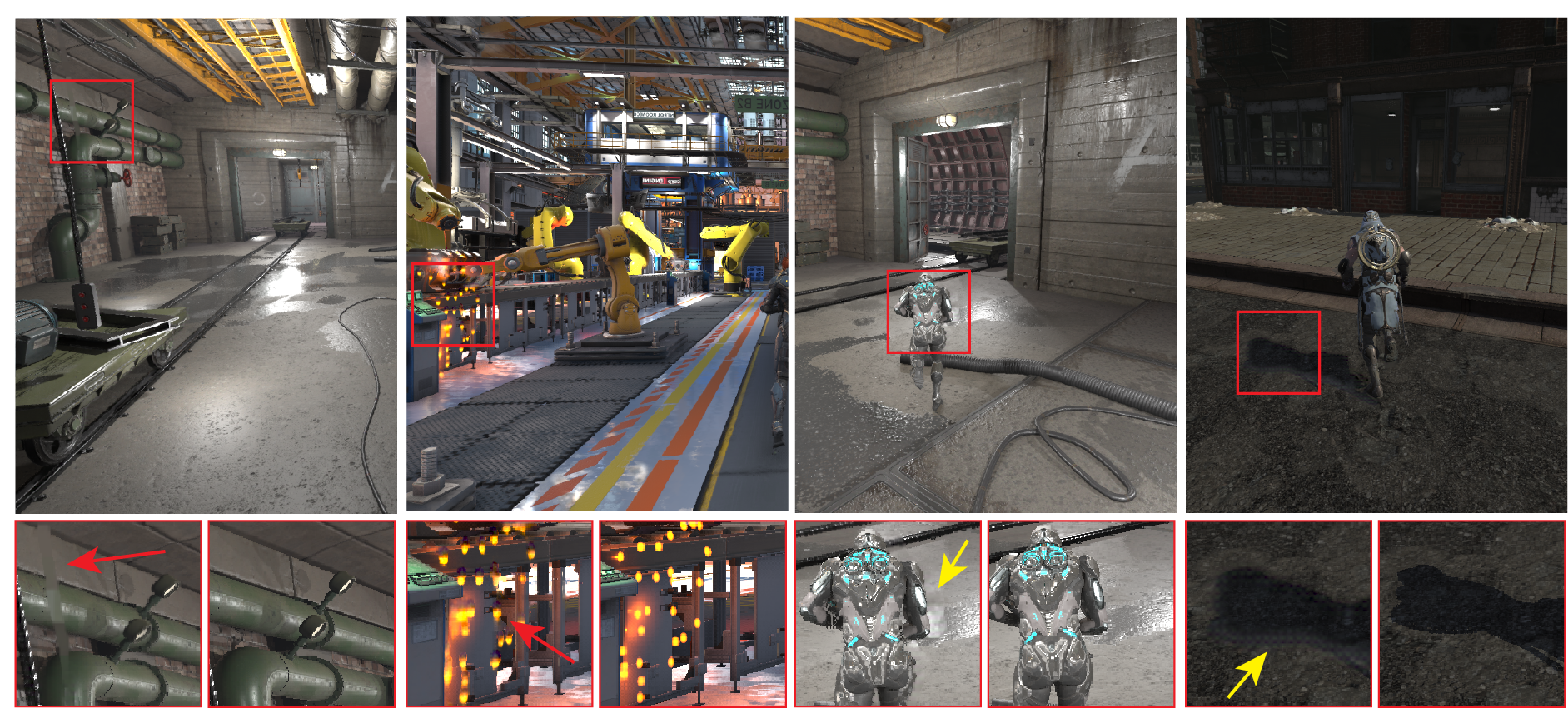}
    \put(1.2, 0.5){\color{white}\small{Ours}}
    \put(13.5, 0.5){\color{white}\small{GT}}
    \put(26.5, 0.4){\color{white}\small{Ours}}
    \put(39, 0.4){\color{white}\small{GT}}
    \put(51, 0.4){\color{white}\small{Ours}}
    \put(63.5, 0.4){\color{white}\small{GT}}
    \put(80.5, 0.4){\color{white}\small{Ours}}
    \put(95, 0.4){\color{white}\small{GT}}
    \end{overpic}
    \caption{Failure cases of our framework includes uncollected disocclusions, effects without depth, shading changes in disocclusions, and imperfect shading correction.}
    \label{fig:failure}
\end{figure}

As noted throughout the paper, our method being G-buffer free extrapolation, has much fewer inputs compared to G-buffer dependent extrapolation (missing G-buffers), and interpolation (missing future frames). Therefore, although with comparable quality overall, our method still has limitations. We analyze them below and show corresponding artifacts in Fig.~\ref{fig:failure}.

\paragraph{Uncollected disocclusions}
Our background collection module tries to find information from previous frames to fill in the disocclusions. However, it fails when the disocclusion areas have never been shown before and are not the out-of-screen areas (Fig.~\ref{fig:failure} the first column).

\paragraph{Effects without depth}
Our framework relies on depth to calculate correct motions and projection. Some effects, including UI and particles, do not have such information, so our framework does not attempt to calculate correct positions in extrapolated frames (Fig.~\ref{fig:failure} the second column). One possible solution could be separating these effects into other passes and combining them with our extrapolated frames.

\paragraph{Shading changes in disocclusions} 
As shown in the third column of Fig.~\ref{fig:failure}, the shading of background collected fragments can be incorrect due to view direction changes, dynamic lighting, and so on. We currently do not specifically train our shading correction network to deal with this and leave it for future work.

\paragraph{Imperfect shading correction} 
Since our method lacks information from G-buffers and future frames compared to the other two types of methods, estimating refined shadings such as shadows are more complicated. As a result, the outcomes of such refined shadings are sometimes blurred (Fig.~\ref{fig:failure}, the fourth column). A better shading correction module is left for future work. 

\section{Conclusion}
\label{sec:conclusion}
We have presented a G-buffer free extrapolation method, GFFE, for low-latency real-time rendering. We addressed three challenges of G-buffer free extrapolation tasks by our designed modules: motion estimation, background collection, adaptive rendering windows and shading correction network. 
 
We evaluated GFFE on diverse scenes and show high quality extrapolation results that demonstrate robustness and generality. The proposed modules provide efficient frame generation without additional latency and extra G-buffers in real-time rendering context. Our framework outperforms G-buffer free extrapolation baselines, and is comparable with frame generation methods including frame interpolation and G-buffer dependent frame extrapolation, with better performance.

In the future, apart from improving the aforementioned limitations, GFFE may be worth exploring in the context of VR/AR and streaming applications. It can also be extended to perform multiple frame extrapolation by passing an extrapolation factor $\alpha$ to the shading correction network to refine the shading motion in different magnitudes to further boost the performance.

\bibliographystyle{ACM-Reference-Format}
\bibliography{paper}

\appendix

\section{Limitation of using G-buffers in extrapolated frames}
As discussed in sec.~\ref{sec:problem_formulation}, G-buffers are not available or become the bottleneck under following cases:
\begin{itemize}
    \item\textbf{Availability:} Some types of G-buffers used in previous extrapolation methods including albedo, roughness and metallic are only available for deferred rendering pipeline. Forward rendering pipeline, which is widely used in smartphone, console and even personal computer platform, doesn't provide such G-buffers.  
    \item\textbf{Complexity:} The generation of G-buffers is the bottleneck in some real-time application. Simulation heavy games require complex simulation process so generating G-buffers is quite time consuming. Besides, some modern game generates high quality G-buffers with low quality shading and then modulate the shading with the G-buffers to render final detailed images, where the generation of G-buffers consumes majority of the time.
    \item\textbf{Memory requirements} Even if the generation of G-buffers is not the bottleneck, it still requires additional memory store them with additional cost in multiple aspects.
\end{itemize}
In these cases, the G-buffer dependent methods~\cite{wu2023extrass, guo2021extranet, wu2023lmv} are limited.

\section{History tracking algorithm}
Here are the details of history collection algorithm. $M^{\text{dyn}}$ is the dynamic mask where $1$ refers to pixels are dynamic fragments.

\begin{algorithm}

    \caption{History tracking with static test}
    \label{alg:motion_fixing}

        \KwData{Pixel position $x$, Current and frame depth \{$D_t$\}, Current motion vector $V_{t\rightarrow t-1}$, Current and previous camera pose \{$C_t$, $C_{t-1}$\}, previous history trajectory $P'$, length of history trajectory $k$}
    
        \KwResult{History trajectory $P$, Dynamic Mask $M^{\text{dyn}}_t$}
        \BlankLine
    
        $p \leftarrow \text{unproject}(x, D_t, C_t)$; \tcp{cur world position}
        
        $\hat{x} \leftarrow \text{project}(p, C_{t-1})$; \tcp{previous position}

        $x' = x + V_{t\rightarrow t-1}[x]$; \tcp{previous position}

        \If(\tcp*[h]{static test}){$\|\hat{x}-x'\|_{2} > \varepsilon$} {
            \For { i = 1 \KwTo k-1}
            {
                $P_i[x] = P'_{i-1}[x']$ \;
            }
            $P_0[x] = p$ \;
            $M^{\text{dyn}}_t[x] \leftarrow 1$ \;
        } 
        \Else(\tcp*[h]{fragment is static}){
            \For {i = 0 \KwTo k-1}
            {
                $P_i[x] = p$ \;
            }
            $M^{\text{dyn}}_t[x] \leftarrow 0$ \;
        }

\end{algorithm}

\section{Adaptive rendering window}
After obtaining current camera pose $C_t$ and estimated next camera pose $\bar{C}_{t+\alpha}$, a virtual plane will be put in front of the current camera along the \texttt{lookat} direction with distance $d$. By calculating the intersections of four corners of camera $C_t$'s view frustum, we get the coordinates of four intersections and corresponding 2D axis-aligned bounding box of them on the plane, denoted as $r = (x_{\text{min}}, y_{\text{min}}, x_{\text{max}}, y_{\text{max}})$. Similarly, we calculate the axis-aligned bounding box of estimated camera $\bar{C}_{t+\alpha}$ on the same virtual plane, denoting as $\bar{r}= (\bar{x}_{\text{min}}, \bar{y}_{\text{min}}, \bar{x}_{\text{max}}, \bar{y}_{\text{max}})$. Then we can calculate the enlarged size of rendering windows based on relative sizes of $r$ and $\bar{r}$.
Assume the original rendering window is the rectangle $(-1,-1, 1, 1)$, the adaptive window of current frame $(u_0, v_0, u_1, v_1)$ is calculated by:
\begin{equation}
\left\{
    \begin{aligned}
        u_0 &= \min(-1, -\bar{x}_{\text{min}} / x_{\text{min}}) \\ 
        v_0 &= \min(-1, -\bar{y}_{\text{min}} / y_{\text{min}}) \\ 
        u_1 &= \max(1,   \bar{x}_{\text{max}} / x_{\text{max}}) \\ 
        v_1 &= \max(1,   \bar{y}_{\text{max}} / y_{\text{max}}) \\ 
    \end{aligned}
\right.
\end{equation}
Note that the virtual plane is put in front of current camera, so the bounding box of current camera on the virtual plane always satisfies $x_{\text{min}} = -x_{\text{max}} < 0$ and $y_{\text{min}} = -y_{\text{max}} < 0$.

\section{Shading Correction Network}
\begin{figure}
    \centering
    \begin{overpic}[width=\columnwidth]{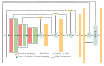}
    \end{overpic}
    \caption{The network structure of shading correction network. The input is down-sampled at first to improve performance.}
    \label{fig:scn_structure}
\end{figure}

\paragraph{Network structure} SCN is a flow-based network with gradually predicted flows to warp intermediate features. The structure of SCN is shown in Fig.~\ref{fig:scn_structure}. The output contains a predicted focus mask and a refined image, and the final output is the blending between the refined image and the input GAE image.

\paragraph{Loss Functions}
To train our SCN, we use the following loss functions to cover various aspects of the output. 

Intermediate feature loss $\mathcal{L}_{f}$ constrains the intermediate features to better align the non-geometric flows from coarse to fine levels. It is defined as:
\begin{equation}
    \mathcal{L}_{f} = \sum\limits_{k=1}^3\mathcal{L}_{cen}(\bar{\phi}_k, \phi_k)
\end{equation}
where $\mathcal{L}_{\text{cen}}$ is the census loss~\cite{meister2018unflow} and $\bar{\phi}_k$ and $\phi_k$ are the intermediate features $k$ of extrapolated frames and ground truth frames from the encoder.

Focus mask loss is the key part of our SCN module to predict a correct focus mask. It is defined as:
\begin{equation}
    \mathcal{L}_{\text{focus}} = \left \|\bar{M}_{\text{focus}} - M_{\text{focus}}\right \|_2
\end{equation}

The reconstruction loss $\mathcal{L}_{\text{recon}}$ is calculated by Charbonnier loss~\cite{charbonnier1994two} between final predicted image and the ground the truth image. The VGG perceptual loss $\mathcal{L}_{\text{vgg}}$ is used to keep the details of extrapolated frames. The final loss function is formulated as
\begin{equation}
    \mathcal{L} = \mathcal{L}_{\text{recon}} + \lambda_f \mathcal{L}_f + \lambda_{\text{focus}} \mathcal{L}_{\text{focus}} + \lambda_{\text{vgg}} \mathcal{L}_{\text{vgg}}
\end{equation}
where we set $\lambda_f = 0.01,\ \lambda_{\text{focus}} = 1.0,\ \lambda_{\text{vgg}} = 0.01$ in our experiments.

\paragraph{Data Preparation}
During the training process, we crop the original images into $256\times 256$ patches to train the network. Since our GAE module provides almost correct geometries, the majority areas of extrapolated frames are correct, which are less useful for training the network. Therefore, we first randomly crop $10^6$ patches from the training dataset, and then sort the crops based on the areas of focus mask $M_{\text{focus}}$. We keep top $15\%$ patches and randomly select other $3\%$ patches for training. We still evaluate on full resolution images during the inference process. All color image in the linear space will be first tone-mapped by $\mu$-Law~\cite{kalantari2017deep} tone-mapper before feeding into the network and the final output will be inverse tone-mapped to the linear space. All losses are calculated in the tone-mapped space.

\paragraph{Training}
We train our model on the cropped dataset with batch size $256$ for $300$ epochs. We use Adam~\cite{kingma2014adam} optimizer with learning rate starting from $10^{-4}$ and gradually decay to $10^{-5}$ during the training. We use PyTorch to implement our network and train it on four NVIDIA A6000 GPUs.

\end{document}